\documentclass[lettersize,journal]{IEEEtran}
\usepackage{amsmath,amsfonts,amssymb}
\usepackage{algorithmic}
\usepackage{algorithm}
\usepackage{array}
\usepackage{booktabs}
\usepackage{subcaption}
\usepackage[group-separator={,}]{siunitx}
\usepackage{textcomp}
\usepackage{stfloats}
\usepackage{url}
\usepackage{verbatim}
\usepackage{graphicx}
\usepackage{cite}
\usepackage{xcolor}
\usepackage{dsfont}
\usepackage{multirow}
\usepackage{float}

\definecolor{arsenic}{rgb}{0.23, 0.27, 0.29}
\definecolor{charcoal}{rgb}{0.21, 0.27, 0.31}
\definecolor{hanblue}{rgb}{0.27, 0.42, 0.81}
\definecolor{blue-ncs}{rgb}{0.0, 0.53, 0.74}
\definecolor{awesome}{rgb}{1.0, 0.13,0.32}
\definecolor{darkgreen}{rgb}{0, .4,0}
\definecolor{purple}{rgb}{.55, .2,.9}

\newcommand{\SBELsimnet}{\textit{$Net_{sim}^{baseline}$}}
\newcommand{\SBELrealnet}{\textit{$Net_{real}$}}

\newcommand{\SBELvarnet}[1]{\textit{$Net^{#1}_{sim}$}}

\newcommand{\SBELcitynet}{\textit{$Net_{city}$}}
\newcommand{\SBELgtanet}{\textit{$Net_{GTA}$}}

\newcommand{\SBELgtavar}[1]{\textit{$Net^{#1}_{GTA}$}}

\newcommand*{\rom}[1]{\expandafter\@slowromancap\romannumeral #1@}

\begin{document}
\title{Camera simulation for robot simulation: \\ 
	how important are various camera model components?}
\author{Asher Elmquist, Radu Serban and Dan Negrut
\thanks{A. Elmquist, R. Serban and D. Negrut are with the University of Wisconsin-Madison. emails: \{amelmquist,serban,negrut\}@wisc.edu }}

\maketitle

\begin{abstract}
	Modeling cameras for the simulation of autonomous robotics is critical for generating synthetic images with appropriate realism to effectively evaluate a perception algorithm in simulation.
	In many cases though, simulated images are produced by traditional rendering techniques that exclude or superficially handle processing steps and aspects encountered in the actual camera pipeline. The purpose of this contribution is to quantify the degree to which the exclusion from the camera model of various image generation steps or aspects affect the sim-to-real gap in robotics. 
	We investigate what happens if one ignores aspects tied to processes from within the physical camera, e.g., lens distortion, noise, and signal processing; scene effects, e.g., lighting and reflection; and rendering quality. 
	The results of the study demonstrate, quantitatively, that large-scale changes to color, scene, and location have far greater impact than model aspects concerned with local, feature-level artifacts. Moreover, we show that these scene-level aspects can stem from lens distortion and signal processing, particularly when considering white-balance and auto-exposure modeling. 
\end{abstract}


\section{Introduction}
\label{sec:intro}

\IEEEPARstart{C}{amera} modeling is an integral component for enabling the simulation of robots as image-based perception is foundational for autonomy. Research into camera modeling and simulation has been steadily improving the visual quality of simulation engines, particularly with the parallel development of photorealistic techniques for games and rendering engines \cite{unityGaming,unrealEngine}. However, autonomy perception tasks are not the primary consideration of these graphics improvements. For simulation of autonomy, the downstream consumers of images are computer-based perception algorithms, not humans. Therefore, photorealism in these applications should be determined by the perception algorithm, not the human eye. 

In recent work \cite{elmquist2022performance}, we introduced a methodology that relied on a perception algorithm, e.g., object detector or semantic segmentation, to quantify the difference between simulated (rendered) and real images. Herein we seek to use the techniques established in \cite{elmquist2022performance} to understand where efforts in modeling cameras can have the largest impact in improving the quality of synthetic data.
As demonstrated here, specific elements of a camera model may be important to a specific algorithm or environment, yet play an insignificant role in a different algorithm or environment. In other words, the proposed methodology is designed to estimate the degree to which specific components of a camera model impact the functional realism of the synthetic data for an algorithm of interest.
Finally, although the approach discussed herein is introduced in conjunction with certain algorithms and environment tests, the methodology outlined generalizes equally well to other algorithms or environments. Against this backdrop, the contributions of this manuscript are:
\begin{enumerate}
	\item a quantitative analysis of the impact of camera sensor model components on the sim-to-real difference;
	\item a demonstration of the relationship between a camera sensor model and the model of the virtual environment;
	\item a demonstration of the importance of scene characteristics and global image effects over micro-level artifacts.
\end{enumerate}

\subsection{Related Work}

There are three main areas of modeling and simulating images covered by this contribution: camera modeling, camera validation, and the study of camera model effects. The most comprehensive and high-fidelity camera modeling and simulation known to the authors comes from the field of computational camera modeling where the focus is the ability to simulate prototype camera designs. While this field focuses more extensively on the camera internals, typically at the expense of simulation speed and versatility, it also provides high-fidelity models of the entire camera system.

These approaches can capture spectral scene modeling, full lens characteristics, precise noise estimation, and complete image processing. Examples of the full pipeline can be found in \cite{liu2019system,retzlaff2017physically,blasinski2018optimizing}. An overview of camera modeling and closed-loop simulation for robotics, and the inclusion of camera component models in such frameworks, can be found in \cite{asherCameraOverview2021}.

Often, to validate such component models, research focuses on pixel- and structural-level comparison between simulated and real images, with comparison performed at a data level. Examples of this methodology are found in \cite{lyu2022validation,grapinet2013optical,gruyer2012modeling} and typically require highly controlled and simple scenes. The approach has been shown to be effective in validation of component level models, and was proposed as a two-stage validation strategy in \cite{durst2022novel}.

While such approaches can be beneficial for understanding low-level models, they place an unnecessary burden on camera simulation, potentially resulting in models that are higher fidelity than needed for perception. Additionally, the controlled nature of the scenes that can be validated makes it difficult to validate the camera models in the complex, uncontrollable scenes where the cameras and associated autonomy algorithms will operate. Therefore, in this work we perform the analysis using an application-focused strategy, where comparison is conducted at the output of a downstream perception algorithm. Our work builds on a recent validation methodology introduced in~\cite{elmquist2022performance} and~\cite{elmquist2022evaluating}.

For the model component evaluation, our work follows a set of inquiries similar to those of Liu et al.~\cite{liu2020neural}, who analyzed camera model parameters to understand their impact on object detection generalization to other datasets. Our work differs in three primary respects:
\begin{itemize}
\item First, we are not interested in the generalization of the perception algorithm, but instead are interested in the predictive nature of the simulation. For example, domain randomization \cite{domainRandomizationAbbeel2017,prakash2019structured,muratore2018domain} has been shown to improve the generalization of trained networks. However, this places undue burden on the perception algorithm to overcome the sim-to-real gap and ignores the inaccuracies of simulation, making simulation less useful when evaluating edge case scenarios.
In simulation, we want to be able to accurately predict real-behavior, even for networks with high specificity. For a given perception algorithm, we seek to produce a simulator that accurately predicts the algorithm's performance in reality, such that lessons learned in closed-loop testing of the algorithm can be reliable and general.
\item Second, we embrace a closed-loop simulation perspective rather than a camera prototyping and system simulation perspective as with ISET/ISET-Auto \cite{farrell2003simulation,blasinski2018optimizing,liu2019system}. Consequently, the rendering pipeline can produce sequences of images in a rapid and efficient manner. To that end, we leverage real-time ray tracing and rendering techniques that inherently impact image quality. Thus, rather than chasing photorealism, we purposely generate images that elicit the same response from a perception algorithm as real images would. 
\item Third, we are less interested in the effect of the model parameters and more interested in the impact of the model itself to understand if inclusion of a particular model feature is, in fact, warranted.
\end{itemize}

Additional work from Liu et al. \cite{liu2019soft} considered the prototyping of new sensor designs to optimize perception algorithm capability and therefore further considered the impact of sensor parameters on the resulting images and subsequent detection algorithm. Where the current work considers parameter effect, we look at model effect to understand the fidelity of simulation needed for closed-loop and online testing of autonomous systems.

Other studies have been conducted on the impact of artifacts on neural networks, by and large the technology of choice for image-based perception, but have focused more on the impact on perception of general artifacts (increasing magnitude of noise, image blur, etc.) rather than the camera models and the impact of their associated phenomena on image-based perception~\cite{dodge2016understanding,zhou2017classification}.

\section{Camera models and indoor lab datasets}
\label{sec:indoor_datasets}

\subsection{Overview and modeling pipeline}

The first application considered herein is a more controlled scenario related to cone detection used by a 1/6th scale autonomous vehicle navigating around a cone-lined course in an indoor lab. This is part of a broader effort in simulation for robotics research \cite{artatk2022}. This task allows for testing in real scenarios which can be subsequently mirrored in simulation. The second application seeks to broaden the scope of the contribution to semantic segmentation in an urban environment, where the scene is complex and cannot be replicated. The models and datasets for this city environment will be discussed in Section~\ref{sec:city_datasets}.

\begin{figure}
	\centering
	\begin{subfigure}[b]{.49\linewidth}
		\centering
		\includegraphics[width=\linewidth]{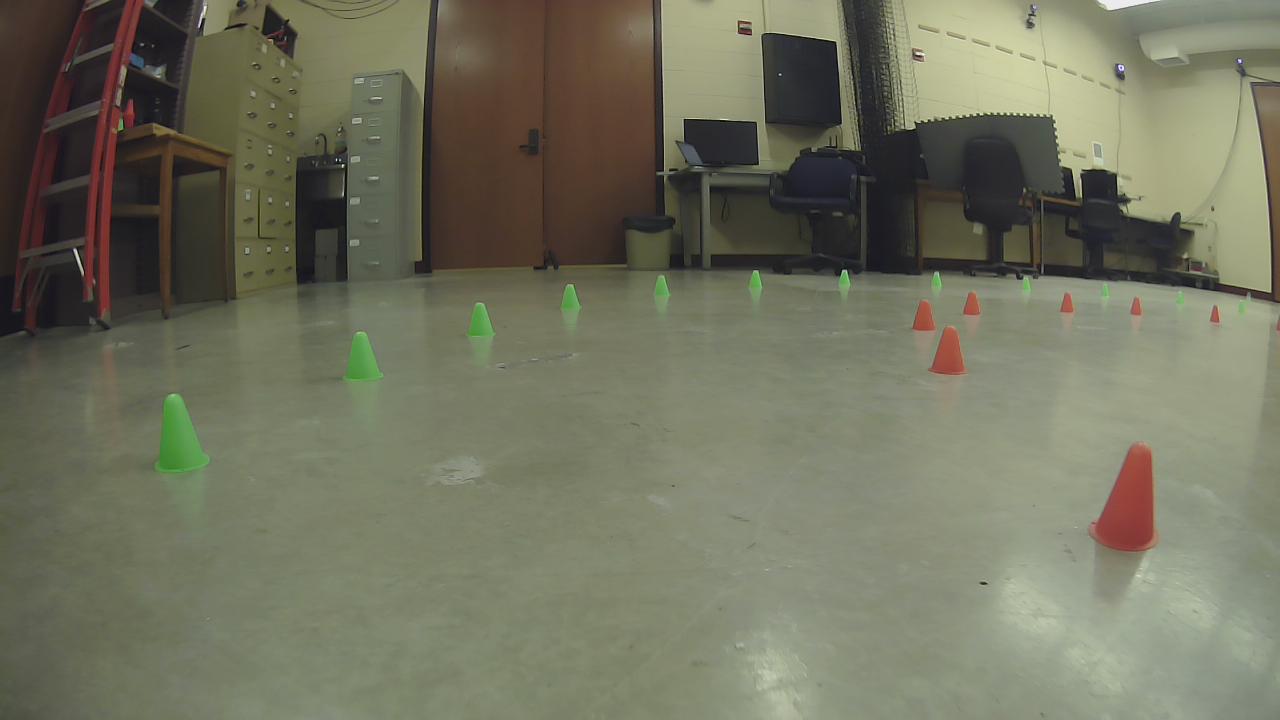}
		\caption{Real}
	\end{subfigure}
	\begin{subfigure}[b]{.49\linewidth}
		\centering
		\includegraphics[width=\linewidth]{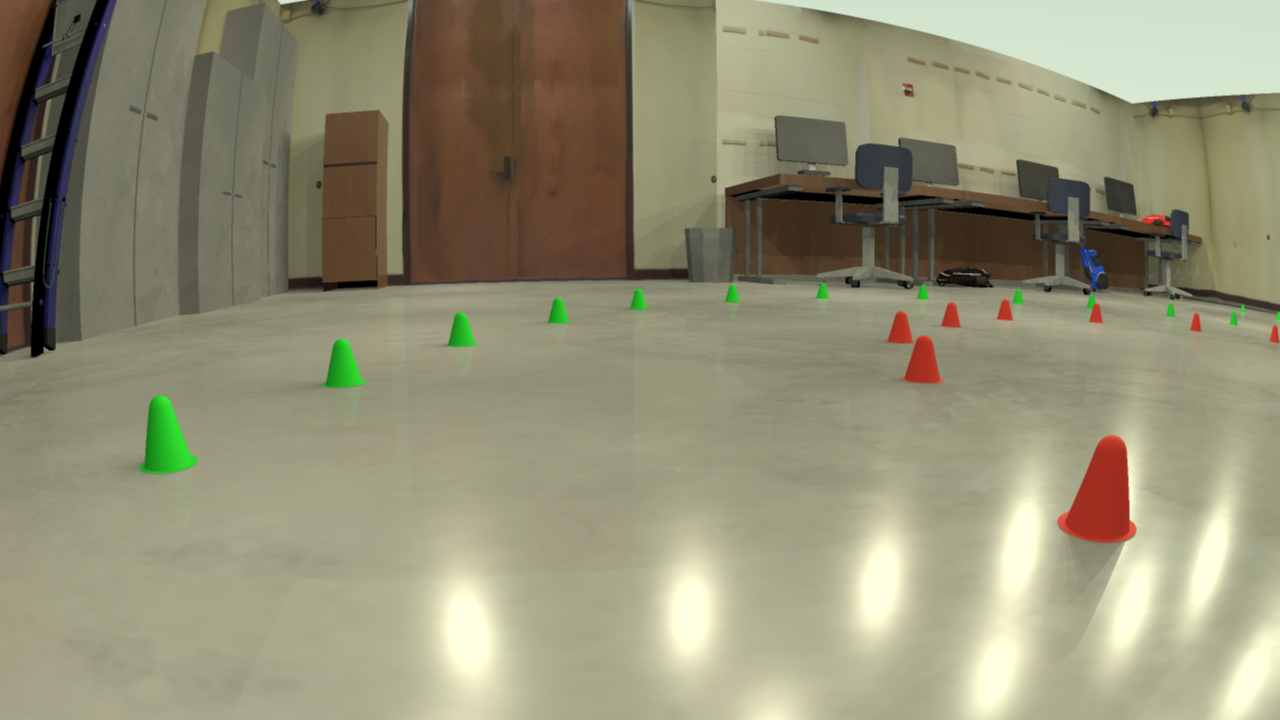}
		\caption{Simulated}
	\end{subfigure}
	\caption{Example real and simulated indoor lab environment with cones. Showing baseline simulation.}
	\label{fig:example_dataset_cones}
\end{figure}

For cone detection, simulation is provided by Chrono \cite{projectChronoWebSite,chronoOverview2016}, with images simulated using Chrono::Sensor \cite{asherSensorSimulation2021}. Chrono::Sensor is a module of Chrono that leverages physically-based rendering and real-time ray tracing to allow modeling of sensors for robotic applications. Beyond camera simulation, Chrono::Sensor supports several other sensors, e.g., lidar, radar, and IMU. Each sensor is characterized by its update rate, lag, noise, and other distortions. Additionally, custom augmentations can be added to the sensor pipeline to model user-defined post-sensor processing. Herein, the 2 mega-pixel camera model is based on an actual USB camera mounted on the 1/6th vehicle. Example real and baseline simulated cone images are provided in Fig.~\ref{fig:example_dataset_cones}.

\begin{figure}
	\centering
	\includegraphics[width=\linewidth]{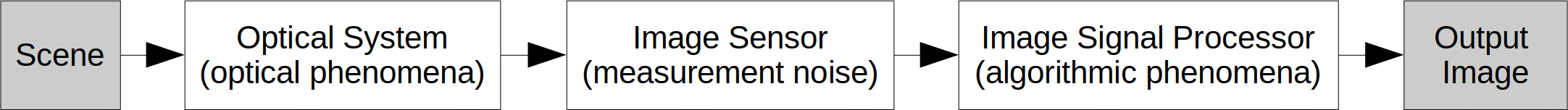}
	\caption{Illustration of the imaging pipeline, showing the progression of light and data through a camera.}
	\label{fig:imaging_pipeline}
\end{figure}

Camera modeling can be broken down into component models of the imaging pipeline illustrated in Fig. \ref{fig:imaging_pipeline}, which are typically implemented within a render engine or graphics pipeline. A detailed description of this pipeline is available in~\cite{farrell2015image}. More information on camera modeling, specifically for autonomous vehicle and robotics, is provided in~\cite{asherCameraOverview2021}. This contribution will look at each section of the imaging pipeline to understand how it impacts the realism of synthetic images, with realism defined from the perspective of an image-based robot perception algorithm.

\subsection{Scene modeling}

In the use of simulated data for training perception algorithms, the virtual environment is understood to play a major role in the realism of simulated data. The domain gap (sim-to-real gap) is a combination of both the appearance and the content differences \cite{prakash2021self}. The content differences can induce bias in training and are largely attributable to more variety and entropy present in real environments. 

Appearance, or how the objects are represented in the data, including texture differences, texture variation, object shapes and colors, scene lighting, etc., can also be significant in testing and training perception algorithms. A question we consider in this paper is the magnitude of the effect of these scene appearance attributes relative to the effect of modeling a camera's internal image acquisition and processing stages. The content distribution's influence over the sim-to-real gap is beyond the scope of this study.

Recent work exists that seeks to optimize the individual models used in the virtual environment such that their renderings are as close as possible to the real object \cite{zhang2022close}. However, we argue that this is intractable on a large scale. For example, specific objects of interest (e.g., cars) could benefit from material optimization techniques, but entire virtual environments, even small ones, typically containing dozens, hundreds, or thousands of objects with many textures, would require time and effort beyond the reasonable scope of simulation efforts. Therefore, we do not consider unbounded modifications to the models used in the environment, but instead consider a small number of visually identifiable characteristics that may go unoptimized in a typical environment. One augmentation is the scene lighting (number of lights). The second is the amount of reflection on the floor (cone reflections in the ground could play significant role in detection).

For the indoor environment, the virtual world was constructed by hand from reference images of the lab which assist in creating realistic textures. The cones were modeled with their color approximated, but not optimized with the sensor pipeline or calibrated to exactly match the cones in the real environment. The real environment was lit by multiple ceiling lights which could not be modeled exactly in Chrono. Considering that lighting directly changes the color and shading of the objects, we consider the impact of how we model these lights.

\begin{figure}
	\centering
	\begin{subfigure}[b]{.470588235\linewidth}
		\centering
		\includegraphics[width=\linewidth]{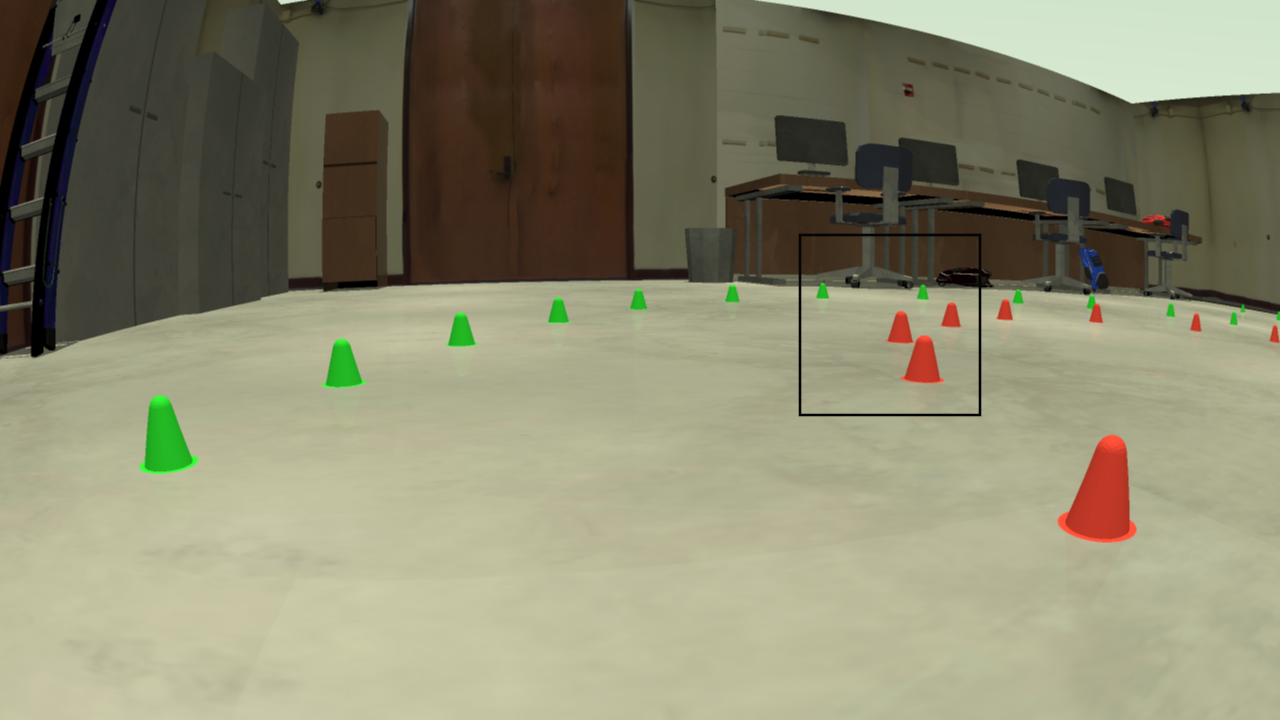}
		\caption{Full augmented frame}
	\end{subfigure}\hskip 0px 
	\begin{subfigure}[b]{0.264705882\linewidth}
		\centering
		\includegraphics[width=\linewidth]{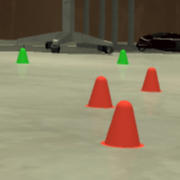}
		\caption{Augmented}
	\end{subfigure}\hskip 0px
	\begin{subfigure}[b]{0.264705882\linewidth}
		\centering
		\includegraphics[width=\linewidth]{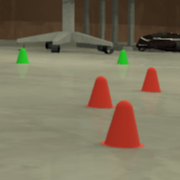}
		\caption{Baseline}
	\end{subfigure}
	\caption{Example of using a single light rather than multiple lights in simulation. Note the difference in shading of the cones.}
	\label{fig:cone:sim_single_light}
\end{figure}

The two light configurations tested are: (i) a single white light source above the environment, which provides even lighting and no reflective highlights; and (ii) 20 white point lights that approximate the beam lighting in the lab and create specular highlights on the floor, which is the configuration for the baseline simulation. Example of the single-light configuration, with a highlighted region of interest is shown in Fig.~\ref{fig:cone:sim_single_light}. For the single light case, the illumination of the cones causes different shading and results in no specular highlights on the floor, even though the floor in both cases has the same reflectance (note the same cone reflections).

\begin{figure}
	\centering
	\begin{subfigure}[b]{.470588235\linewidth}
		\centering
		\includegraphics[width=\linewidth]{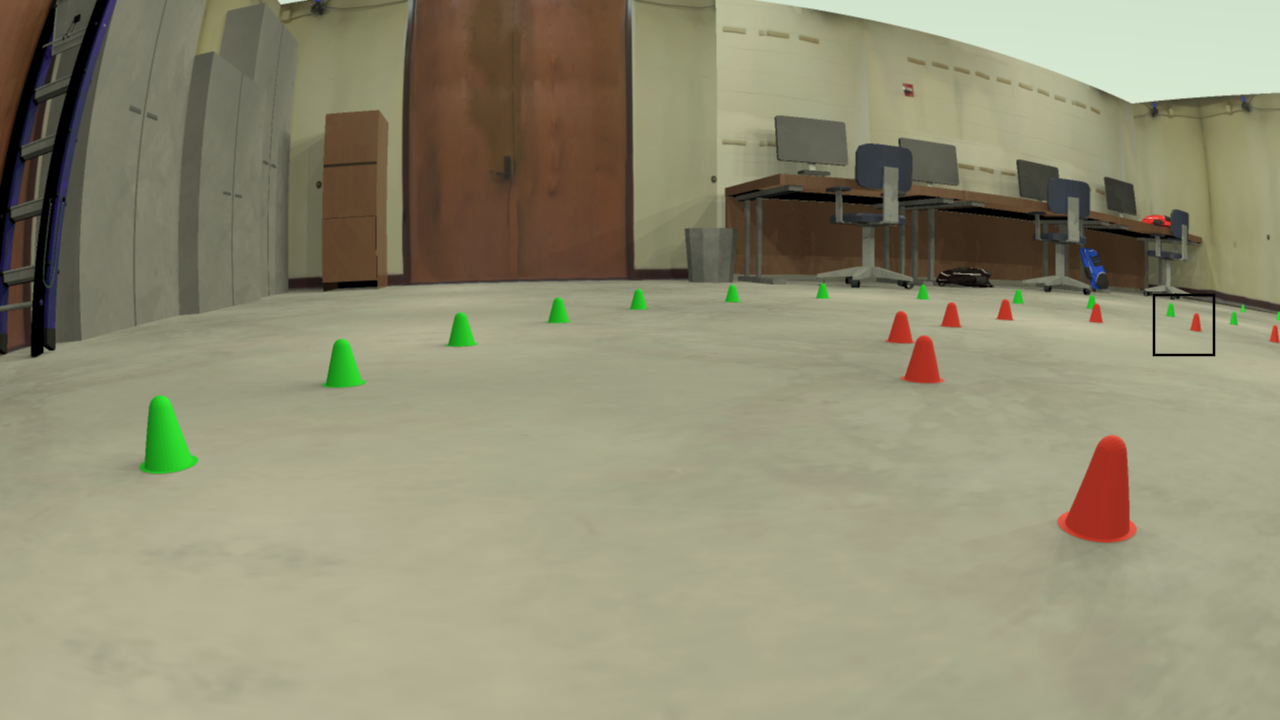}
		\caption{Full augmented frame}
	\end{subfigure}\hskip 0px 
	\begin{subfigure}[b]{0.264705882\linewidth}
		\centering
		\includegraphics[width=\linewidth]{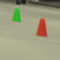}
		\caption{Augmented}
	\end{subfigure}\hskip 0px
	\begin{subfigure}[b]{0.264705882\linewidth}
		\centering
		\includegraphics[width=\linewidth]{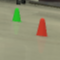}
		\caption{Baseline}
	\end{subfigure}
	\caption{Example of making the ground non-reflective in the simulation.}
	\label{fig:cone:sim_noreflect}
\end{figure}

Since reflections were prominent in the real images (see Fig.~\ref{fig:example_dataset_cones}), the floor material parameters were calibrated to produce visually similar reflections for the baseline simulation. A simulation variant with no floor reflections is also considered to evaluate the impact of the reflections on the perception algorithm. See Fig.~\ref{fig:cone:sim_noreflect} for an example of the difference with no reflection on the floor. To remove the reflections of the cones, the floor's material parameters were modified such that no reflections are possible.

\subsection{Data acquisition: image quality}
\begin{figure}
	\centering
	\begin{subfigure}[b]{.470588235\linewidth}
		\centering
		\includegraphics[width=\linewidth]{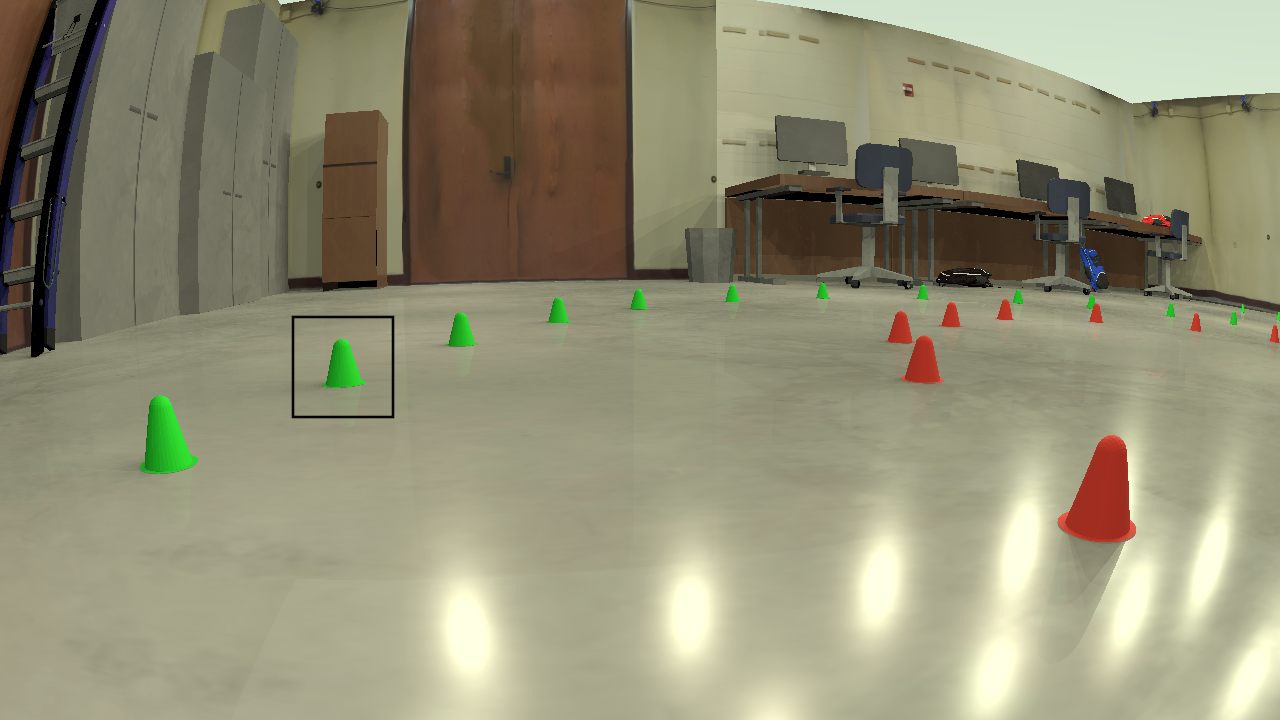}
		\caption{Full augmented frame}
	\end{subfigure}\hskip 0px 
	\begin{subfigure}[b]{0.264705882\linewidth}
		\centering
		\includegraphics[width=\linewidth]{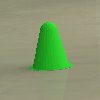}
		\caption{Augmented}
		\label{fig:cone:sim_single_sample:roi}
	\end{subfigure}\hskip 0px
	\begin{subfigure}[b]{0.264705882\linewidth}
		\centering
		\includegraphics[width=\linewidth]{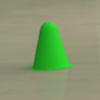}
		\caption{Baseline}
	\end{subfigure}
	\caption{Example using a single ray per pixel to generate the synthetic images. Note the hard edges of the single sample image.}
	\label{fig:cone:sim_single_sample}
\end{figure}

Rendering can be performed via two general algorithms: rasterization and ray tracing. As the simulation in this contribution uses ray tracing, we will focus on the number of rays that are used, per pixel, to sample the scene to determine the final colors in the image. When a single ray is used, each pixel can only include the color from a single object. For example, near the edge of a cone, a ray either hits the cone or goes past, so the edges of object will be hard and zoomed-in views of those obstacles may appear pixelated (see Fig.~\ref{fig:cone:sim_single_sample:roi}). When additional rays are used, these can be used to super sample the pixel in multiple locations, producing an anti-aliased image and a better approximation of soft edges for real images. To compare the impact of this image quality, we generated an augmented dataset using one sample (ray) per pixel (spp), with the baseline using four spp. An example image rendered with one spp, with a highlighted region of interest, alongside the same region of interest from the baseline simulation is shown in Fig.~\ref{fig:cone:sim_single_sample}.

\subsection{Optical model: lens distortion}

Lens systems play a key role in the accumulation of light on the image sensor, potentially distorting the rays to accommodate for additional light and wider fields of view. The real camera which is to be modeled has a wide field of view and generates visible distortions near the edges. In this work, we use three different lens models. First, a pinhole camera is used, a common non-distorted model employed in game rendering. Second, we analyze a low-fidelity, single-parameter model based on the field of view (FOV) model \cite{sturm2011camera}, which can capture some distortion but makes simplistic assumptions about the geometry of the lens. Lastly, we consider a radial model commonly used in computer vision which serves as the highest-fidelity model considered; with well calibrated parameters, this was found to give distortions accurate to within 1\% of the true distortion of the actual camera. 

To set the field of view of the pinhole model, we used the effective field of view such that the horizontal axis of the images captures the same extent of the scene as the real camera. The FOV model was parameterized by the calibrated focal distance. Lastly, the radial model was calibrated using the MATLAB image calibration toolbox \cite{matlab-cameraCalibrator}.

\begin{figure}
	\centering
	\begin{subfigure}[b]{.32\linewidth}
		\includegraphics[width=\linewidth]{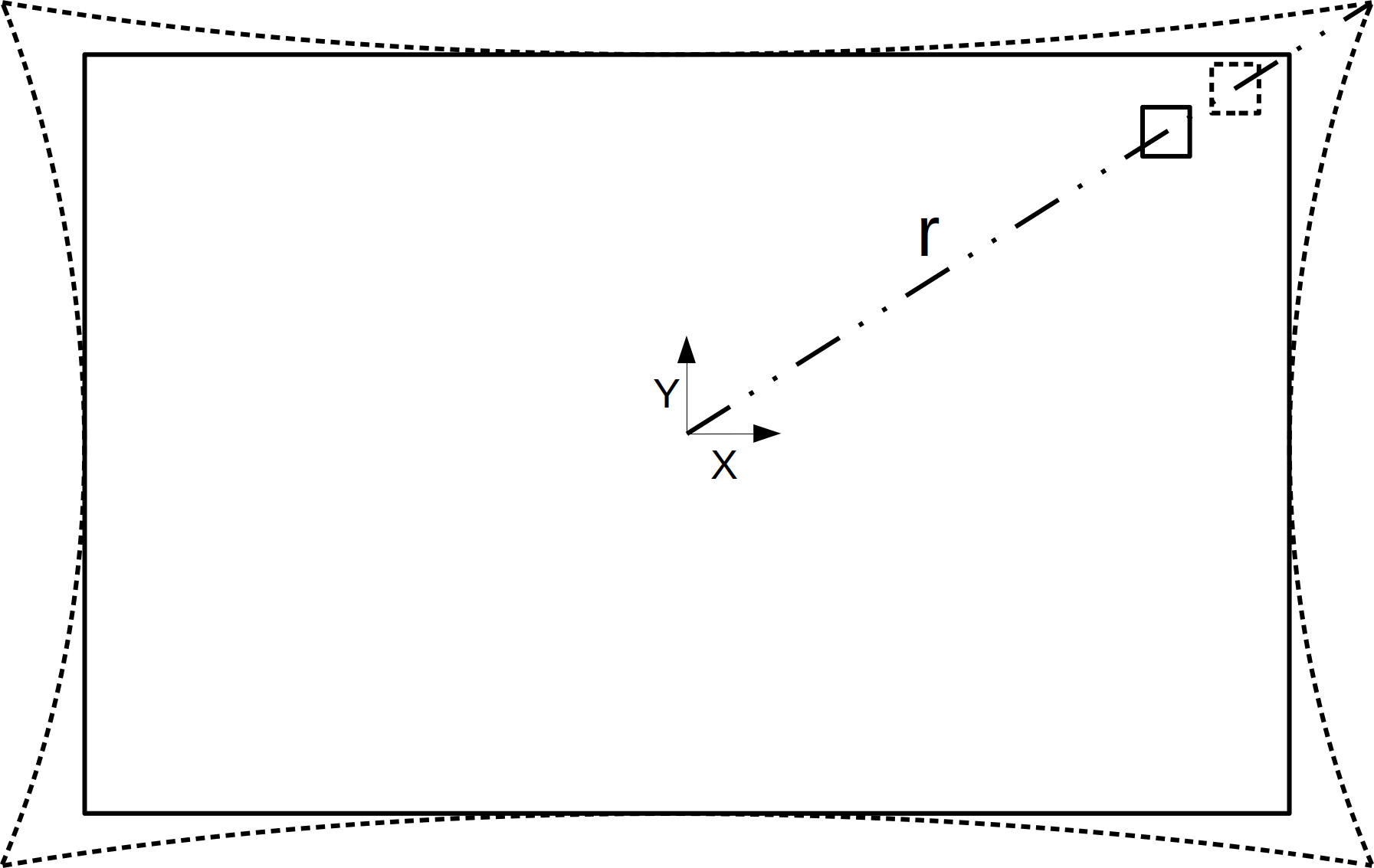}
		\caption{}
	\end{subfigure}
	\begin{subfigure}[b]{.32\linewidth}
		\includegraphics[width=\linewidth]{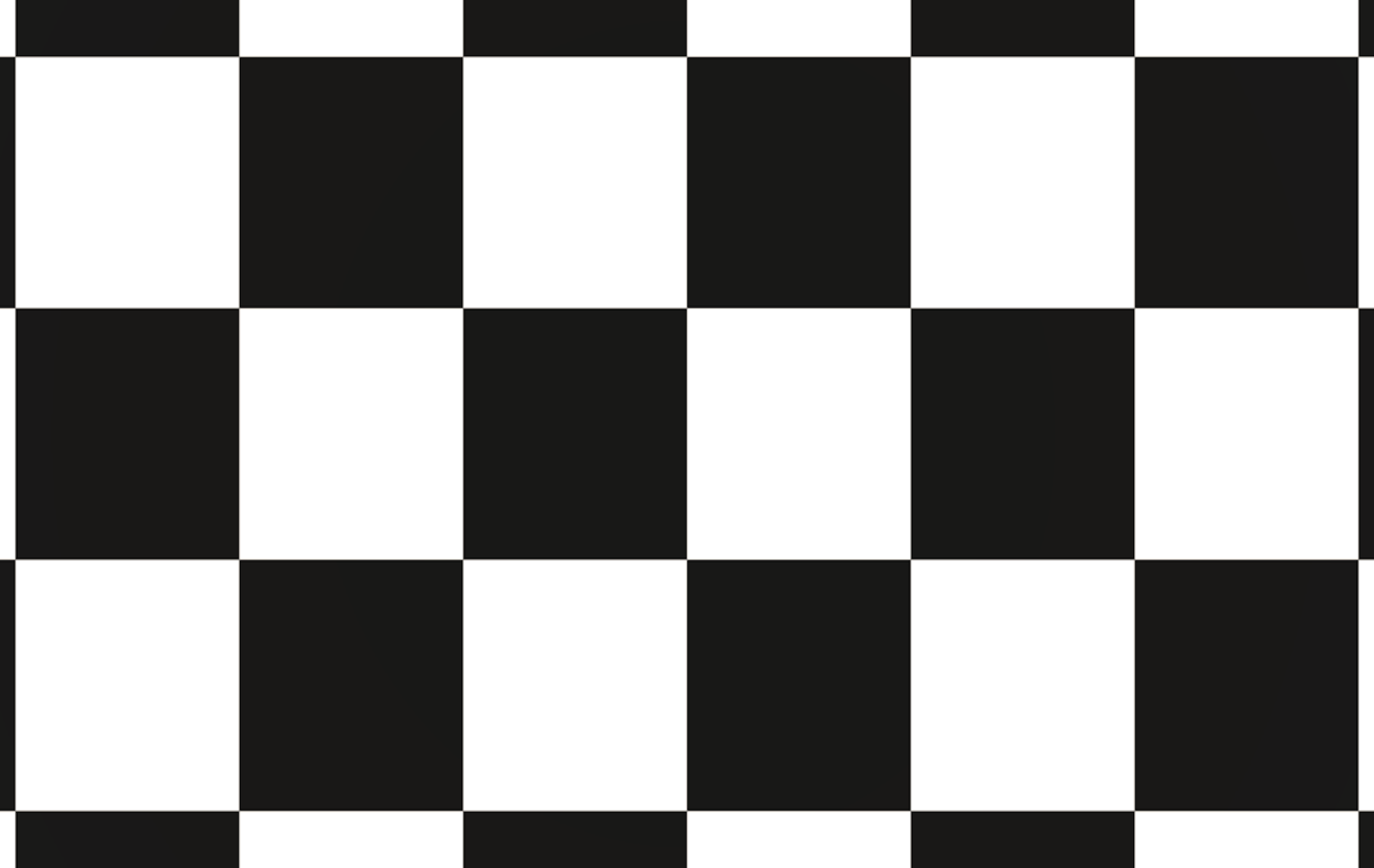}
		\caption{}
	\end{subfigure}
	\begin{subfigure}[b]{.32\linewidth}
		\includegraphics[width=\linewidth]{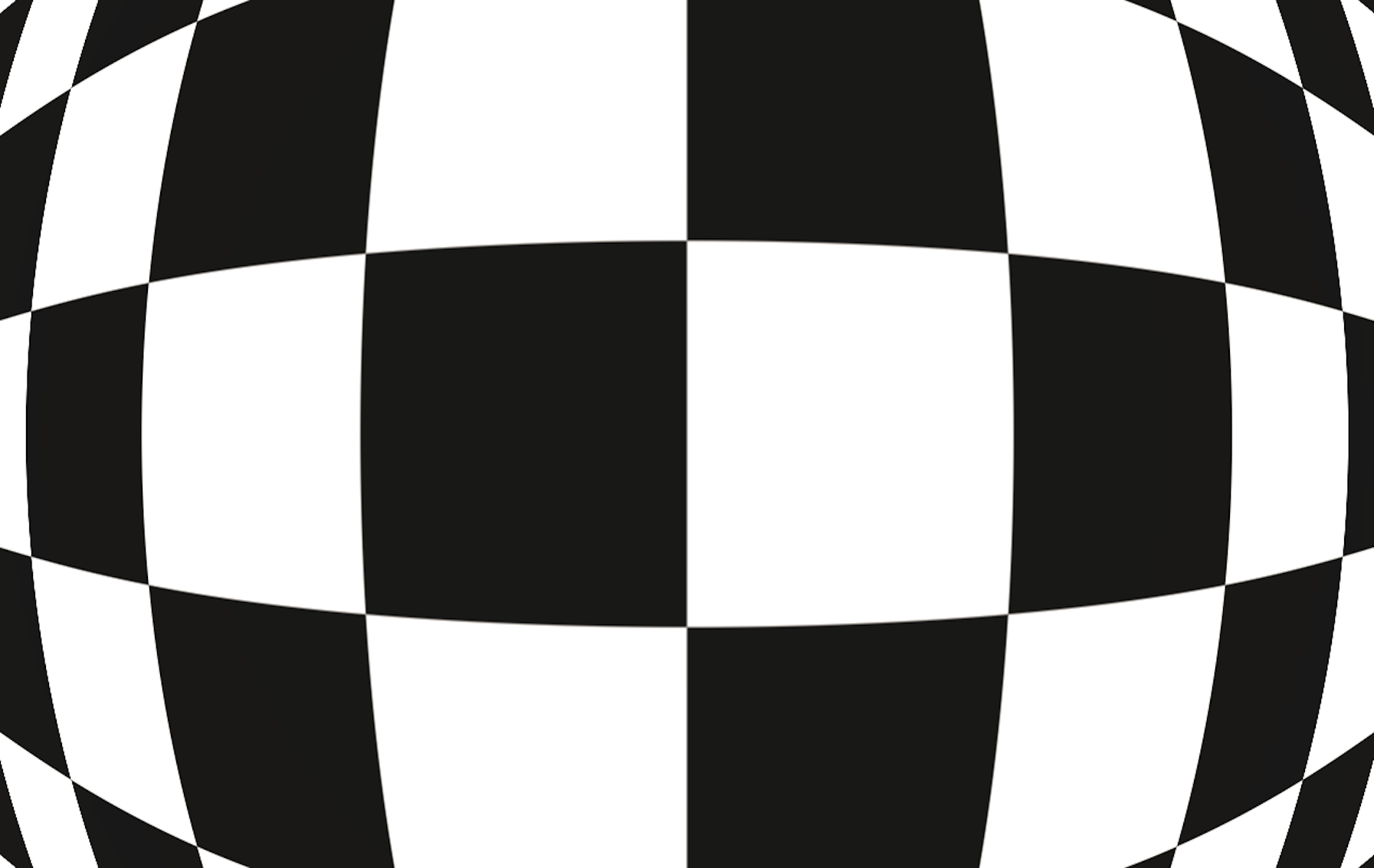}
		\caption{}
	\end{subfigure}
	\caption{a) radial warping distorts pixels radially from the optical center. b) checkerboard rendered with pinhole model (no distortion). c) checkerboard rendered with a radial model. Effective horizontal field of view is maintained.}
	\label{fig:models:lens_distortion}
\end{figure}

\begin{figure}
	\centering
	\begin{subfigure}[b]{.470588235\linewidth}
		\centering
		\includegraphics[width=\linewidth]{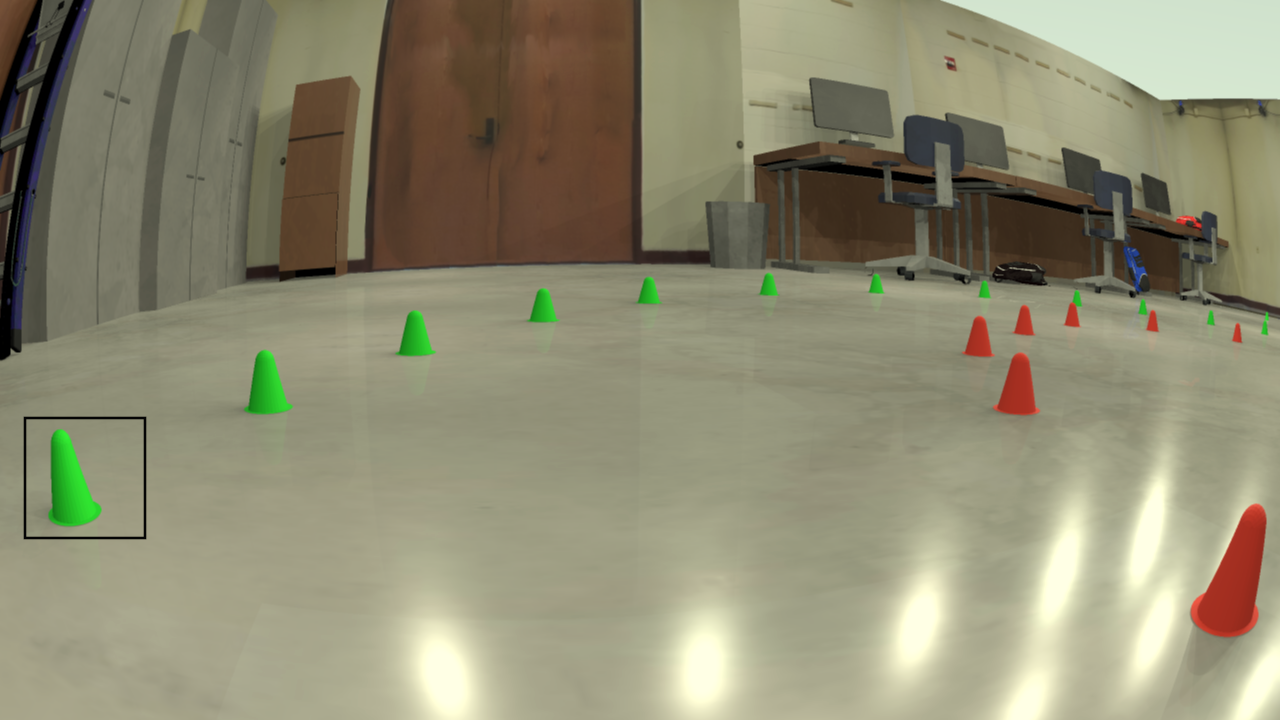}
		\caption{Full augmented frame}
	\end{subfigure}\hskip 0px 
	\begin{subfigure}[b]{0.264705882\linewidth}
		\centering
		\includegraphics[width=\linewidth]{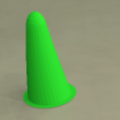}
		\caption{Augmented}
	\end{subfigure}\hskip 0px
	\begin{subfigure}[b]{0.264705882\linewidth}
		\centering
		\includegraphics[width=\linewidth]{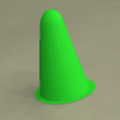}
		\caption{Baseline}
	\end{subfigure}
	\caption{Example using a lower-fidelity, single-parameter distortion model in the simulation. Not the slight difference in cones near the edges of the image.}
	\label{fig:cone:sim_fov_lens}
\end{figure}

\begin{figure}
	\centering
	\begin{subfigure}[b]{.470588235\linewidth}
		\centering
		\includegraphics[width=\linewidth]{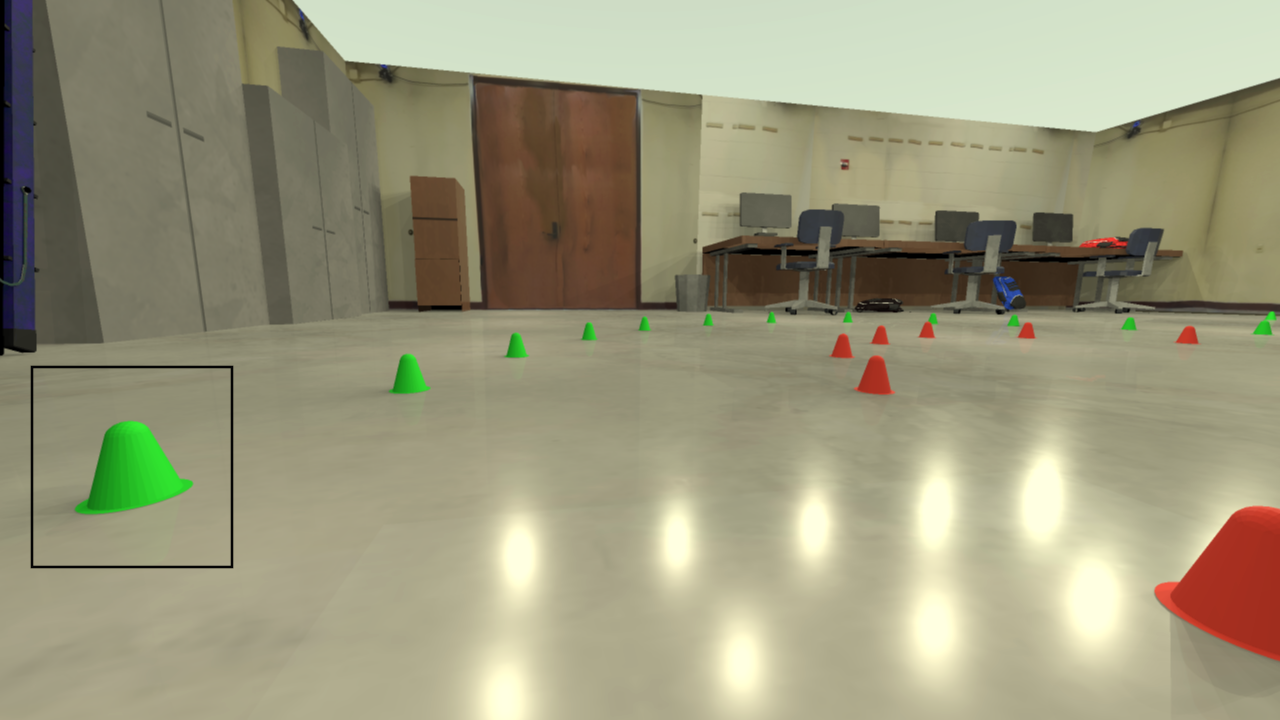}
		\caption{Full augmented frame}
	\end{subfigure}\hskip 0px 
	\begin{subfigure}[b]{0.264705882\linewidth}
		\centering
		\includegraphics[width=\linewidth]{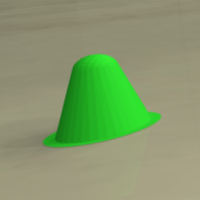}
		\caption{Augmented}
	\end{subfigure}\hskip 0px
	\begin{subfigure}[b]{0.264705882\linewidth}
		\centering
		\includegraphics[width=\linewidth]{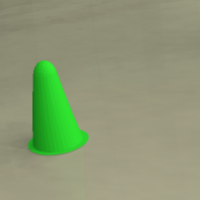}
		\caption{Baseline}
	\end{subfigure}
	\caption{Example using a pinhole camera, and ignoring lens distortion in the camera. Not the wider appearance of cones near the edges of the image relative to the baseline case.}
	\label{fig:cone:sim_pinhole_lens}
\end{figure}

While the baseline simulation used a radial distortion model, we seek here to understand whether using simpler models or ignoring lens distortion would provide similar results. Therefore, we consider two alternatives: (a) no distortion via a pinhole model, and (b) wide angle distortion via the FOV model. Examples that highlight the difference between these models can be seen in Figs.~\ref{fig:cone:sim_fov_lens} and \ref{fig:cone:sim_pinhole_lens}. The primary visual difference is the aspect ratio of cones near the edges of the images.

\subsection{Image sensor: noise}
\label{sec:models:noise}

The image sensor's noise can be modeled using several approaches which have been found to effectively approximate noise and the image sensor level ~\cite{hasinoff2010noise,standard20101288,liu2008automatic}. Noise modeling is not straightforward, since onboard processing of the raw image can significantly distort the pixel- and chromatically-independent noise through algorithms such as demosaicing, denoising, deblurring, gamma correction, and compression. Therefore, these existing noise models may not precisely capture the noise in the final image. 

\begin{figure}
	\centering
	\begin{subfigure}[b]{.470588235\linewidth}
		\centering
		\includegraphics[width=\linewidth]{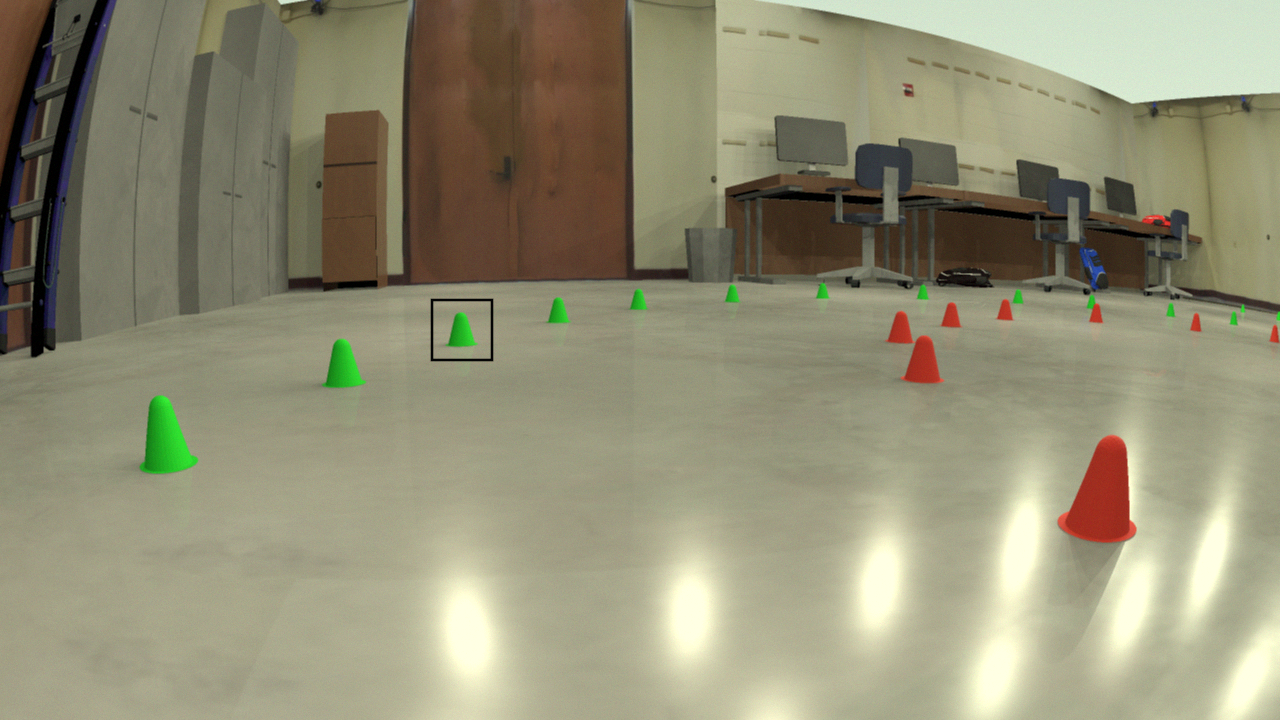}
		\caption{Full augmented frame}
	\end{subfigure}\hskip 0px 
	\begin{subfigure}[b]{0.264705882\linewidth}
		\centering
		\includegraphics[width=\linewidth]{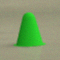}
		\caption{Augmented}
	\end{subfigure}\hskip 0px
	\begin{subfigure}[b]{0.264705882\linewidth}
		\centering
		\includegraphics[width=\linewidth]{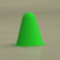}
		\caption{Baseline}
	\end{subfigure}
	\caption{Example of adding white Gaussian noise ($\sigma=0.01$) to the simulated cone images.}
	\label{fig:cone:sim_awgn}
\end{figure}

Due to images being rendered in sRGB space, estimated noise on the real images being low, and the scene radiance being unknown, higher fidelity models were excluded from comparison in favor of understanding the more general impact of noise on training and testing perception. The noise model considered here is an additive white Gaussian noise (AWGN). AWGN is a simplistic noise model applied to make the data less idealistic and often only used when the noise is of unknown origin. The estimated noise of the real images had standard deviation less than half the image precision. This indicates that, between image processing and downsizing, very little noise remained and as such noise was excluded from the baseline simulation. For the study, we seek to understand the impact of introducing noise in the simulation, for both low- and high-fidelity models. To understand the impact of this change, AWGN was applied with $\sigma=0.01$ to images with color 0-1.  For an 8-bit-per-channel image, this is approximately 2.5$\times$ the image precision. We will consider and discuss the choice of magnitude further when analyzing the impact in Section~\ref{sec:results}. Example of a noisy image can be seen in Fig.~\ref{fig:cone:sim_awgn}.

\subsection{Image signal processing (ISP): demosaicing, exposure, and color balance}
\begin{figure}
	\centering
	\includegraphics[width=.85\linewidth]{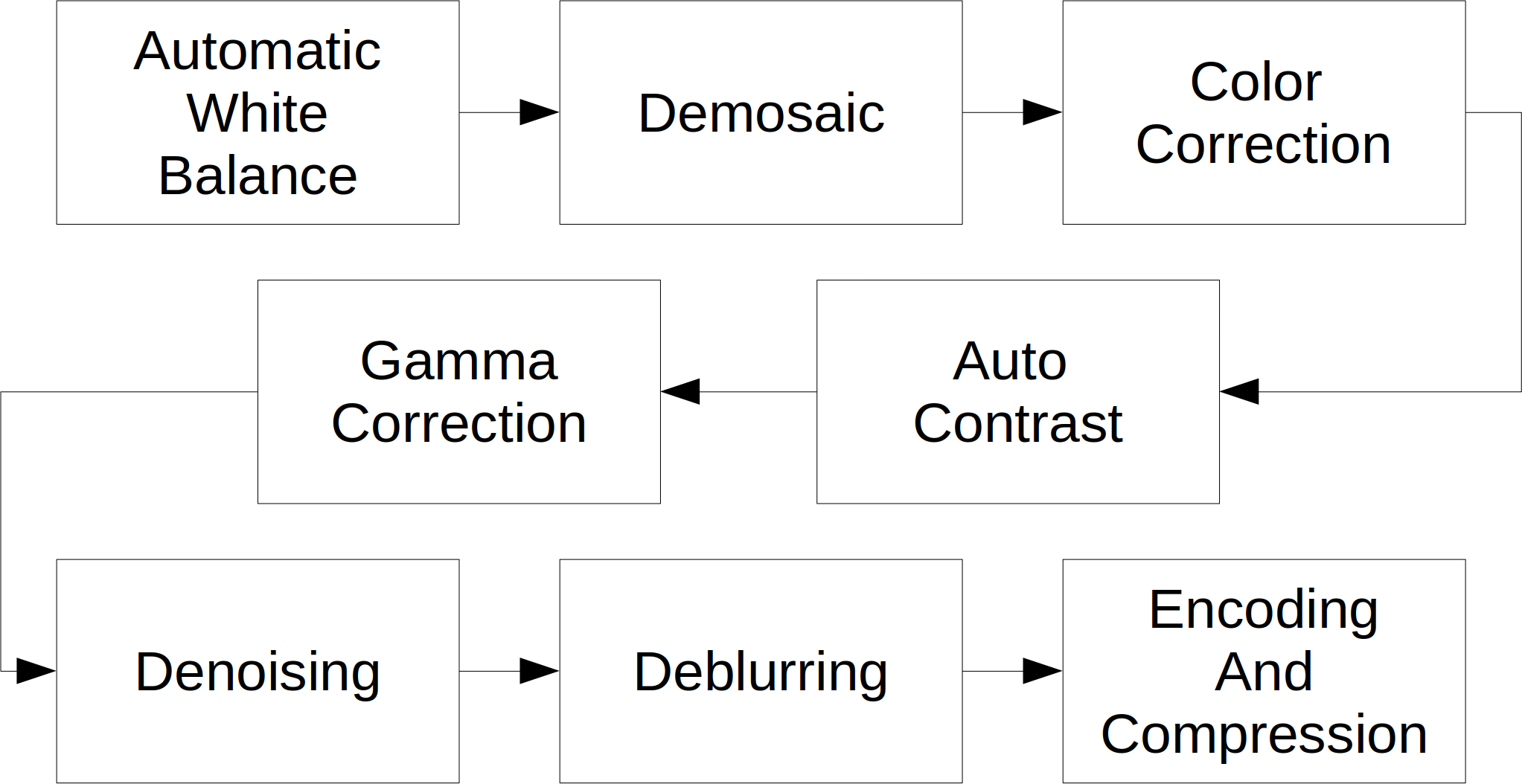}
	\caption{Example ISP \cite{choi2014implementation} showing a collection of algorithms that can alter the quality of an image.}
	\label{fig:models:isp}
\end{figure}

Cameras capture intensity over an array of light-sensitive pixels, called the image sensor. To capture distinct red, green, and blue values that make up the three channels of an RGB image, the sensor leverages a mosaic of color filters which provide the intensity of light at a given location for a single channel. A common filter pattern is the red-green-green-blue (RGGB) Bayer pattern, which we will consider here. Other patterns, such as RCCC (red-clear-clear-clear), also exist for dedicated purposes in automotive sensing. The image collected by this pattern is the raw image which is then converted to an RGB image through a demosaicing process that interpolates the missing RGB values from nearby pixels~\cite{malvar2004high}. The image can be further processed by the ISP to denoise, deblur, color correct, white balance, or otherwise modify the image. A set of example post-sensor processing operations is provided in Fig.~\ref{fig:models:isp}.

These post-sensor operations are often overlooked when simulating robotics using off-the-shelf game or rendering engines, thus failing to capture potentially significant augmentations to the final image. Here, we consider the effect of three post-sensor algorithms: demosaicing, exposure, and white balancing.

\begin{figure}
	\centering
	\includegraphics[width=.5\linewidth]{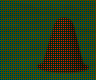}
	\caption{Image viewed with Bayer RGGB pattern}
	\label{fig:implementation:demosaic}
\end{figure}

\begin{figure}
	\centering
	\begin{subfigure}[b]{.470588235\linewidth}
		\centering
		\includegraphics[width=\linewidth]{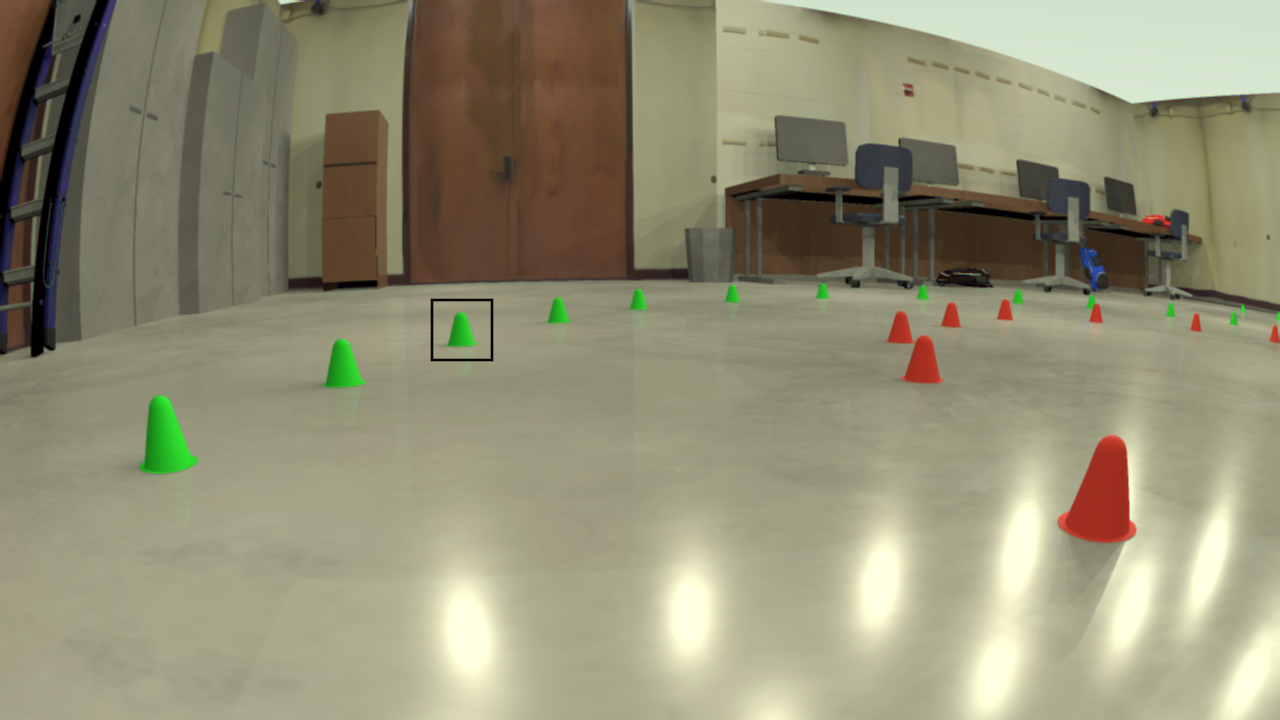}
		\caption{Full augmented frame}
	\end{subfigure}\hskip 0px 
	\begin{subfigure}[b]{0.264705882\linewidth}
		\centering
		\includegraphics[width=\linewidth]{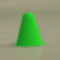}
		\caption{Augmented}
	\end{subfigure}\hskip 0px
	\begin{subfigure}[b]{0.264705882\linewidth}
		\centering
		\includegraphics[width=\linewidth]{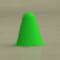}
		\caption{Baseline}
	\end{subfigure}
	\caption{Example of modeling the raw-to-RGB demosaicing converstion in the camera.}
	\label{fig:cone:sim_demoisaiced}
\end{figure}

Since real images are often captured with a Bayer pattern, demosaicing can soften object edges and lead to slight color bleeding. For the indoor environment, we simulate images as RGB, then convert to raw by sampling the color in the Bayer pattern, see Fig.~\ref{fig:implementation:demosaic}. We then use OpenCV~\cite{opencv_library} and an edge aware demosaicing algorithm to convert back to RGB. An example of the resulting demosaiced image is shown in Fig.~\ref{fig:cone:sim_demoisaiced}.

\begin{figure}
	\centering
	\begin{subfigure}[b]{.470588235\linewidth}
		\centering
		\includegraphics[width=\linewidth]{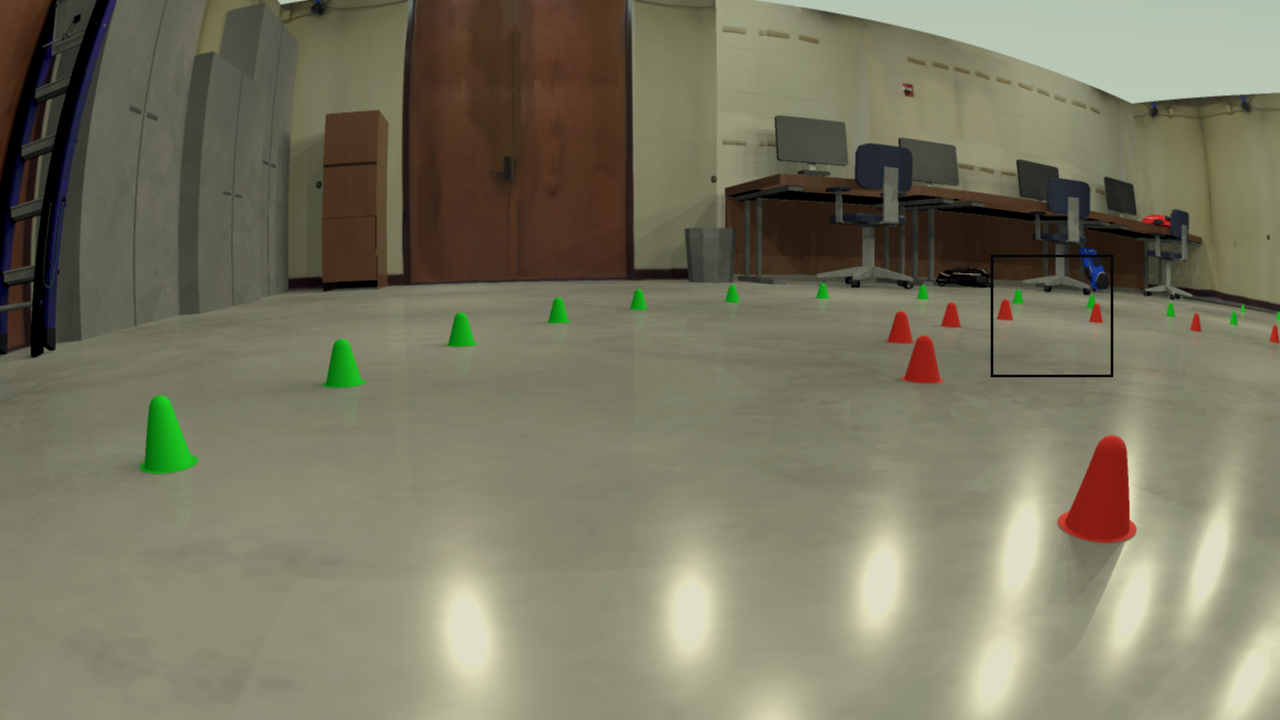}
		\caption{Full augmented frame}
	\end{subfigure}\hskip 0px 
	\begin{subfigure}[b]{0.264705882\linewidth}
		\centering
		\includegraphics[width=\linewidth]{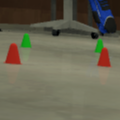}
		\caption{Augmented}
	\end{subfigure}\hskip 0px
	\begin{subfigure}[b]{0.264705882\linewidth}
		\centering
		\includegraphics[width=\linewidth]{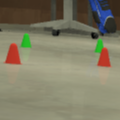}
		\caption{Baseline}
	\end{subfigure}
	\caption{Example of matching the exposure levels of the real and simulated images.}
	\label{fig:cone:sim_exposure}
\end{figure}

\begin{figure}
	\centering
	\begin{subfigure}[b]{0.49\linewidth}
		\centering
		\includegraphics[width=\linewidth]{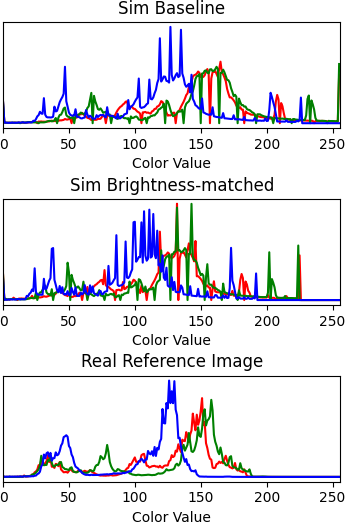}
		\caption{Brightness matching}
		\label{fig:implementation:colorcorrect_histogram}
	\end{subfigure}
	\begin{subfigure}[b]{.49\linewidth}
		\centering
		\includegraphics[width=\linewidth]{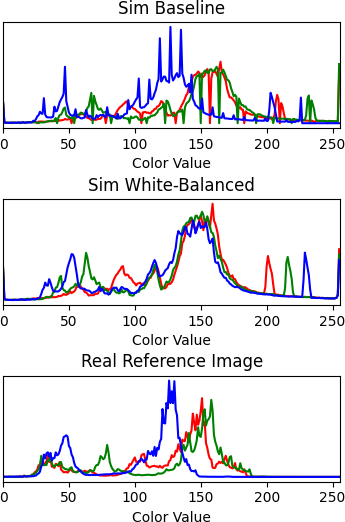}
		\caption{White-balanced}
		\label{fig:implementation:ECWB_histogram}
	\end{subfigure}
	\caption{Color histograms for baseline, augmented, and real images when color augmentation models are applied.}
	\label{fig:implementation:cone_histograms}
\end{figure}

To understand the impact of exposure, we note the difference in the mean and standard deviations of brightness in simulated vs. real images. We then augment the simulated images so that their mean and standard deviations match the mean of a calibration set consisting of real images. We do not fully match histograms since the variety of colors present in the simulated and real images are not equivalent. The visual effect is to slightly darken the simulated images, as seen in Fig.~\ref{fig:cone:sim_exposure}. A comparison of the histograms between baseline simulation, augmented simulation, and real image are provided in Fig.~\ref{fig:implementation:colorcorrect_histogram}. Other auto-contrast/brightening algorithms exist \cite{maini2010comprehensive,vishwakarma2012color}, but would be challenging to use for matching the mean characteristics of the real data.

Additionally, we consider a variation of the exposure model for the case where the mean and standard deviation of the color channels had been perfectly reconstructed for each image. This is possible since the metric dataset has sim-real images pairs. For the detector training dataset, we consider a stochastic pairing between the sim and real detector training sets. This results in matching not only the exposure, but also the variation of exposure in the datasets.

\begin{figure}
	\centering
	\begin{subfigure}[b]{.470588235\linewidth}
		\centering
		\includegraphics[width=\linewidth]{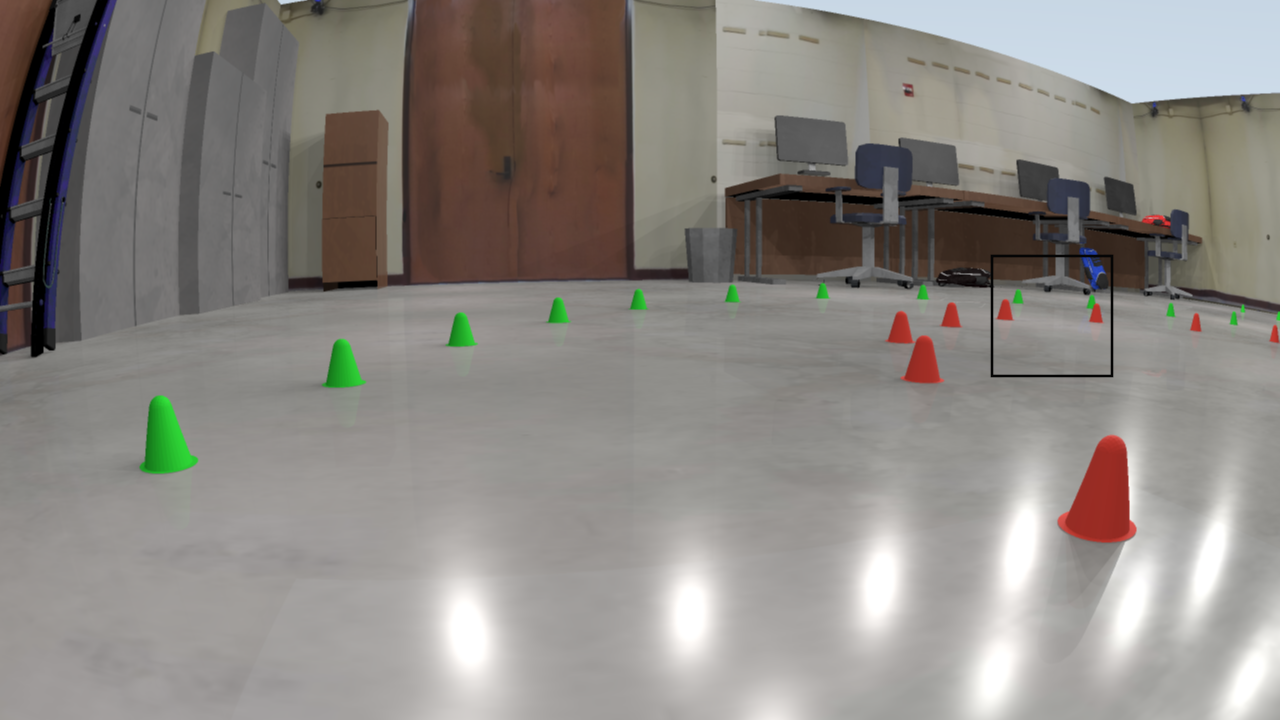}
		\caption{Full augmented frame}
	\end{subfigure}\hskip 0px 
	\begin{subfigure}[b]{0.264705882\linewidth}
		\centering
		\includegraphics[width=\linewidth]{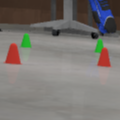}
		\caption{Augmented}
	\end{subfigure}\hskip 0px
	\begin{subfigure}[b]{0.264705882\linewidth}
		\centering
		\includegraphics[width=\linewidth]{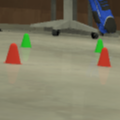}
		\caption{Baseline}
	\end{subfigure}
	\caption{Example of including white-balancing in the simulation.}
	\label{fig:cone:colorbalanced}
\end{figure}

Using a traditional computer graphics pipeline, which typically does not consider pixel sensitivities, results in rendered data with correct color balancing and white balancing. During real image dataset capture, it was noted that the images were not white balanced, either because no white balancing was enabled on the camera or because the white balancing algorithm was insufficient. As this significantly changed the overall color of the images, we modeled it in simulation by adjusting the color gains to match the mean white balance gains from a real calibration set. The algorithm followed a gray-world white balancing method, but distorted the image away from a gray world, to the mean of the real images. To understand the impact of this decision, we compare a correctly white-balanced simulation with the baseline simulation. The correctly white balanced version is the simulation without the color-distortion model. An example balanced image is provided in Fig.~\ref{fig:cone:colorbalanced}.

\section{City environment datasets}
\label{sec:city_datasets}

\begin{figure}
	\centering
	\begin{subfigure}[b]{.49\linewidth}
		\centering
		\includegraphics[width=\linewidth]{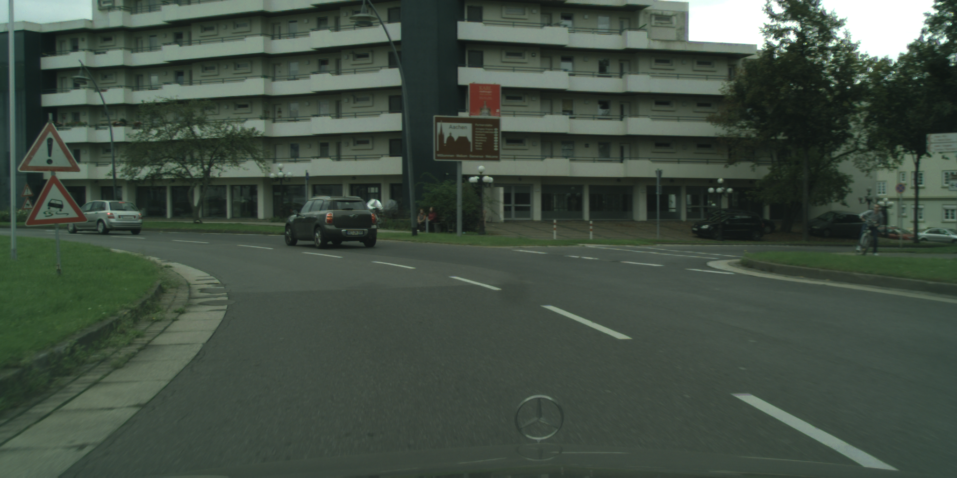}
		\caption{Real}
	\end{subfigure}
	\begin{subfigure}[b]{.49\linewidth}
		\centering
		\includegraphics[width=\linewidth]{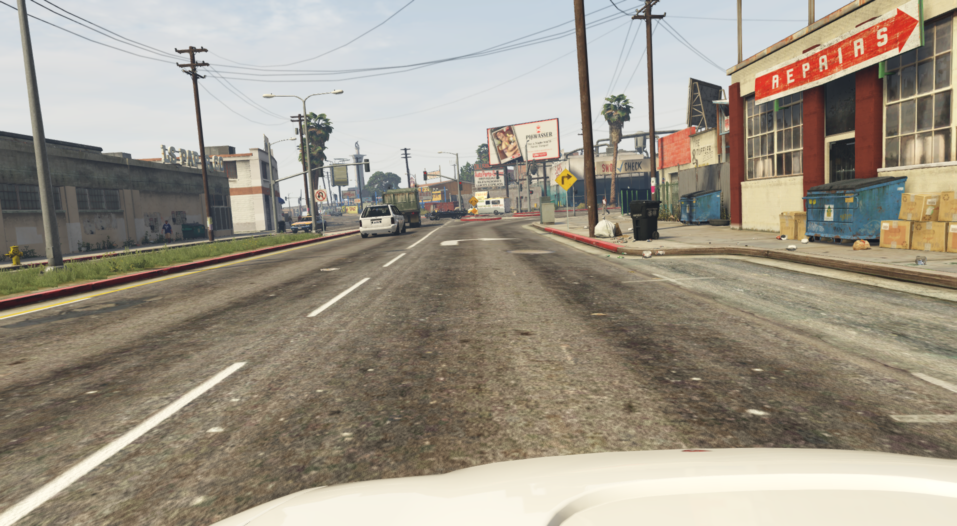}
		\caption{Simulated}
	\end{subfigure}
	\caption{Example real (Cityscapes) and simulated (GTAV) city environments. }
	\label{fig:example_dataset_city}
\end{figure}

For semantic segmentation, Cityscapes~\cite{cordts2016cityscapes} was used as the real data, and the GTAV dataset~\cite{richter2016playing} was used as simulation. GTAV was chosen as the model of Cityscapes due to the consistent labels and the qualitatively similar content. Example Cityscapes (real) and GTAV (simulated) images are provided in Fig.~\ref{fig:example_dataset_city}. The GTAV data with no augmentations is treated as the baseline. The only modification made to the data is to reduce both Cityscapes and GTAV images to a width of 957 to match the width of reference GAN-enhanced data provided by~\cite{richter2021enhancing}, which can serve as a reference point for the effect of a GAN approach to modeling the real data. For each augmentation that was applied in post-process to GTAV, train/validation/test sets were generated.

\noindent\textbf{Scene modeling.} Since the GTAV dataset is provided directly, we have no control and cannot modify aspects of the environment such as lighting, materials, etc. These effects are therefore ignored in this study. For post-processing GTAV images and enhancing the image's scene appearance, see work related to generative adversarial networks such as~\cite{richter2021enhancing} and a related analysis of this algorithm on realism for semantic segmentation~\cite{elmquist2022evaluating}.

\noindent\textbf{Data acquisition: image quality.} Similar to scene modeling, we ignore the impact of rendering quality as we cannot appropriately alter the data in the post-process stage.

\begin{figure}
	\centering
	\begin{subfigure}[b]{.476356396\linewidth}
		\centering
		\includegraphics[width=\linewidth]{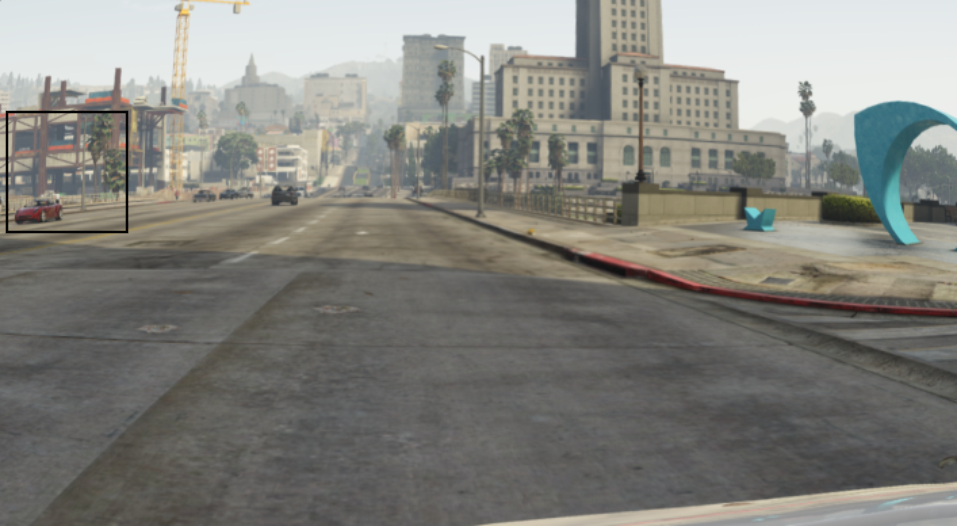}
		\caption{Full augmented frame}
	\end{subfigure}\hskip 0px 
	\begin{subfigure}[b]{0.261821802\linewidth}
		\centering
		\includegraphics[width=\linewidth]{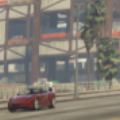}
		\caption{Augmented}
	\end{subfigure}\hskip 0px
	\begin{subfigure}[b]{0.261821802\linewidth}
		\centering
		\includegraphics[width=\linewidth]{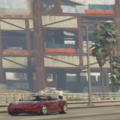}
		\caption{Baseline}
	\end{subfigure}
	\caption{Example of a GTAV image post-processed with a wide-angle lens model.}
	\label{fig:gta:radial}
\end{figure}

\noindent\textbf{Optical model: lens distortion.} While it is unknown if a lens-distortion model was applied during the rendering of GTAV, we make the assumption here that a pinhole camera was used. This is very common in game engines and this assumption is supported by the fact that objects near the edge of the images appeared undistorted. Therefore, we seek to understand the impact of lens distortion by applying the same radial model as used in the indoor simulation to the GTAV images as a post-process operation, preserving the image size. An example of a disorted GTAV image can be seen in Fig.~\ref{fig:gta:radial}.

\begin{figure}
	\centering
	\begin{subfigure}[b]{.476356396\linewidth}
		\centering
		\includegraphics[width=\linewidth]{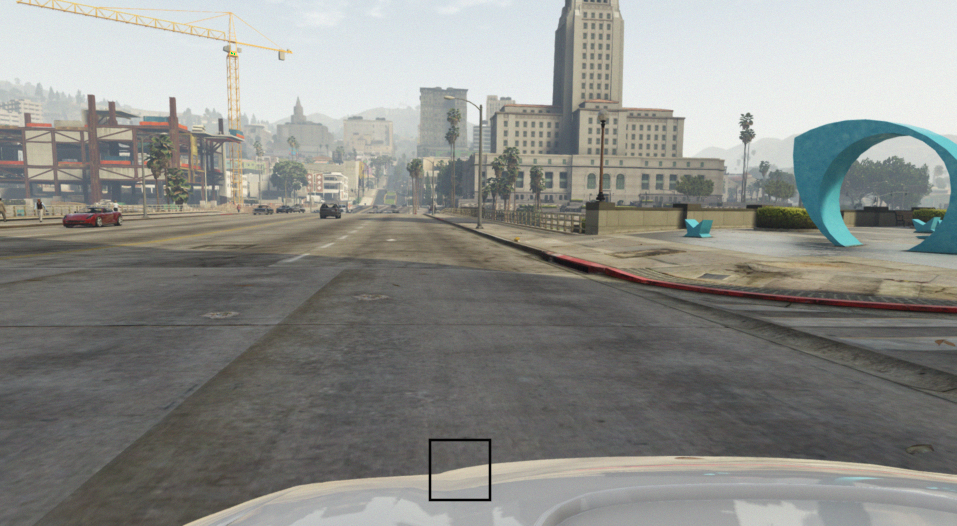}
		\caption{Full augmented frame}
	\end{subfigure}\hskip 0px 
	\begin{subfigure}[b]{0.261821802\linewidth}
		\centering
		\includegraphics[width=\linewidth]{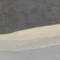}
		\caption{Augmented}
	\end{subfigure}\hskip 0px
	\begin{subfigure}[b]{0.261821802\linewidth}
		\centering
		\includegraphics[width=\linewidth]{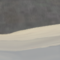}
		\caption{Baseline}
	\end{subfigure}
	\caption{Example of adding white Gaussian noise ($\sigma=0.01$) to the GTAV images.}
	\label{fig:gta:awgn}
\end{figure}

\noindent\textbf{Image sensor: noise.} As with real cones, Cityscapes and GTAV had low estimated noise level, lower than half the image precision. It is likely that GTAV had little to no noise initially, and between any onboard processing for Cityscapes and the downsizing done herein, noise was minimal. However, we still seek to understand the degree to which realism would be affected if noise were included. Therefore, similar to the cone dataset, AWGN was applied with $\sigma=0.01$. Example of the noised images for GTAV can be seen in Fig.~\ref{fig:gta:awgn}.

\begin{figure}
	\centering
	\begin{subfigure}[b]{.476356396\linewidth}
		\centering
		\includegraphics[width=\linewidth]{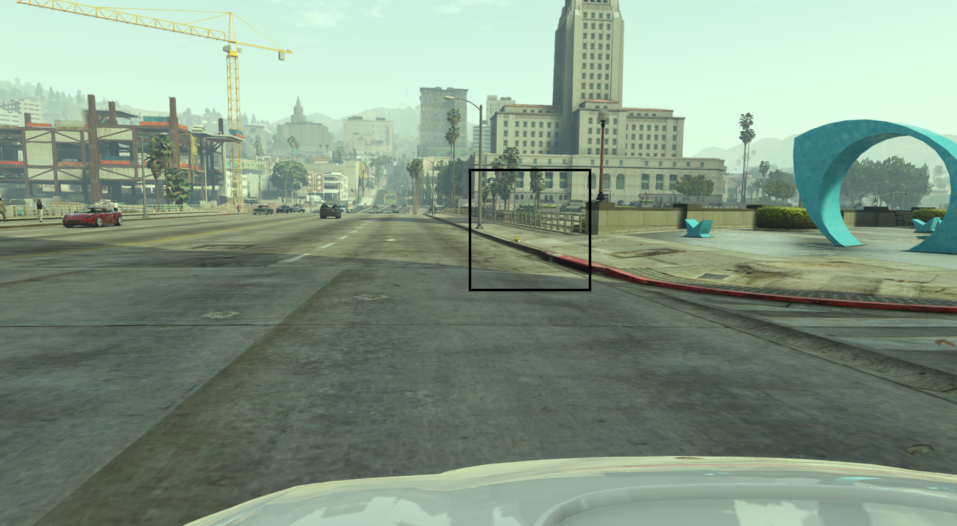}
		\caption{Full augmented frame}
	\end{subfigure}\hskip 0px 
	\begin{subfigure}[b]{0.261821802\linewidth}
		\centering
		\includegraphics[width=\linewidth]{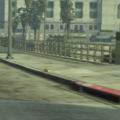}
		\caption{Augmented}
	\end{subfigure}\hskip 0px
	\begin{subfigure}[b]{0.261821802\linewidth}
		\centering
		\includegraphics[width=\linewidth]{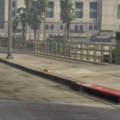}
		\caption{Baseline}
	\end{subfigure}
	\caption{Example of white-balance matching between GTAV and Cityscapes datasets.}
	\label{fig:gta:wb}
\end{figure}

\begin{figure}
	\centering
	\begin{subfigure}[b]{.476356396\linewidth}
		\centering
		\includegraphics[width=\linewidth]{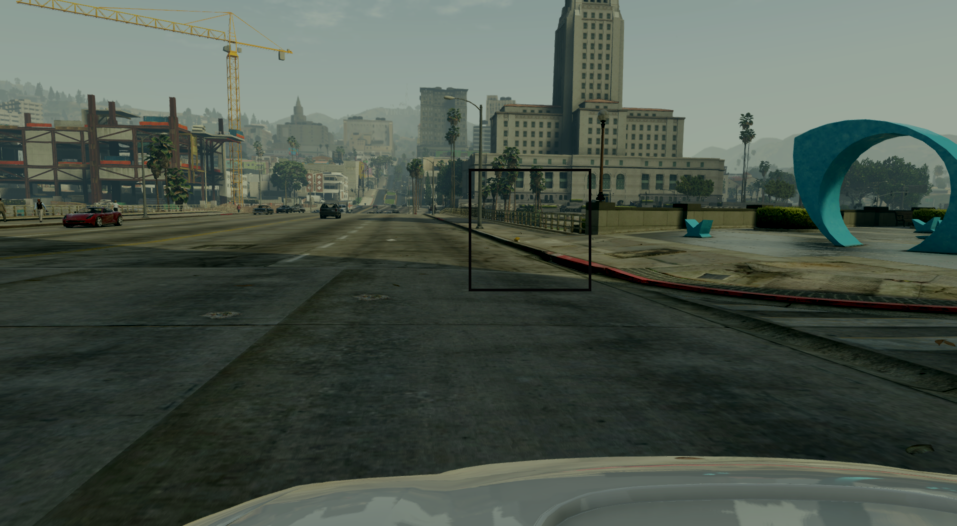}
		\caption{Full augmented frame}
	\end{subfigure}\hskip 0px 
	\begin{subfigure}[b]{0.261821802\linewidth}
		\centering
		\includegraphics[width=\linewidth]{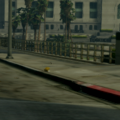}
		\caption{Augmented}
	\end{subfigure}\hskip 0px
	\begin{subfigure}[b]{0.261821802\linewidth}
		\centering
		\includegraphics[width=\linewidth]{images/city_datasets/fullcolordistort_roi_ref.png}
		\caption{Baseline}
	\end{subfigure}
	\caption{Example of matching the color balance and exposure levels between GTAV and Cityscapes datasets.}
	\label{fig:gta:ECWB}
\end{figure}

\noindent\textbf{ISP: demosaicing, exposure, and color balance.} Because of the aforementioned downsizing, we ignore demosaicing as an augmentation of GTAV as it would only impact fine-grain detail. The two post-sensor augmentations we consider for GTAV are white balancing and brightness adjustment. We model these changes in the same way as for cones images, where white-balancing brings the GTAV data into the same off-white mode as a Cityscapes calibration set. For brightness matching, we again adjust the GTAV images to match the mean and standard deviation of a set of Cityscapes calibration images. See examples of white balance matching in Fig.~\ref{fig:gta:wb} and additional brightness matching in Fig.~\ref{fig:gta:ECWB}.

\begin{figure}
	\centering
	\begin{subfigure}[b]{.49\linewidth}
		\includegraphics[width=\linewidth]{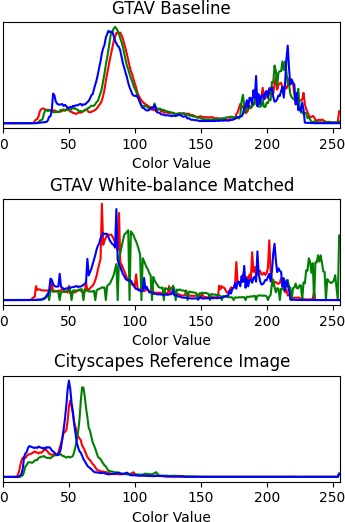}
		\caption{White-balance matching}
		\label{fig:gta:colorcorrect_histogram}
	\end{subfigure}
	\begin{subfigure}[b]{0.49\linewidth}
		\centering
		\includegraphics[width=\linewidth]{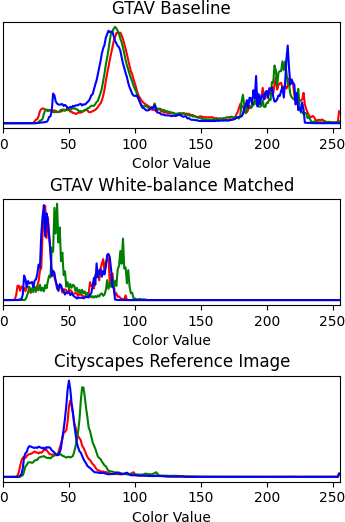}
		\caption{Brightness-matching}
		\label{fig:gta:ECWB_histogram}
	\end{subfigure}
	\caption{Example color histograms for augmented-color GTAV images.}
	\label{fig:gta:histograms}
\end{figure}

To demonstrate the color distortion, Fig.~\ref{fig:gta:colorcorrect_histogram} shows the histograms of GTAV, Cityscapes, and the white-balance matched image. Figure~\ref{fig:gta:ECWB_histogram} shows the histograms for a GTAV, Cityscapes, and brightness-matched image.

\section{Perception and comparison algorithms}
\label{sec:networks}

Given many variants of the indoor lab simulation, we seek to understand their differences from the perspective of a perception algorithm. The perception algorithm of choice for the indoor lab is object detection of the two cone classes. The object detector is YOLOv5 Nano~\cite{jocher2020yolov5}. For each version of the data (real, baseline, and each variant) a version of the detector is trained using data exclusively from its respective domain. These networks will be used as a way to judge similarity of the datasets and the magnitude of the augmentations. Each network was trained to convergence on its own validation set. All hyperparameters were consistent for all networks. From here, these object detector networks are referred to as: {\SBELrealnet} for the real dataset, {\SBELsimnet} for the baseline dataset, and {\SBELvarnet{augmentation}} for variants.

For the city environment, the perception algorithm of choice is a semantic segmentation network, leveraging the work from~\cite{tao2020semanticsegmentation} and the accompanying library~\cite{nvidiaSemanticSegmentation}. This algorithm was demonstrated to be able to train highly accurate networks on Cityscapes data. For each city dataset (Cityscapes, GTAV, and GTAV variants) we train a semantic segmentation network exclusively on that dataset, until convergence on a same-domain validation set. All hyperparameters were consistent for all networks. From here, these semantic segmentation networks are referred to as: {\SBELcitynet} for the Cityscapes dataset, {\SBELgtanet} for the GTAV baseline, and {\SBELgtavar{augmentation}} for the GTAV variants.

The methodology we use here to compare the datasets follows the contextualized performance difference proposed in~\cite{elmquist2022performance} and further generalized in~\cite{elmquist2022evaluating}. This methodology evaluates the difference in performance distribution for batches of similar content between the datasets. Herein, performance is defined as the efficacy of the perception algorithm (object detection or semantic segmentation).

\section{Results}
\label{sec:results}

\subsection{Indoor lab}

\begin{figure}
	\centering
	\includegraphics[width=\linewidth]{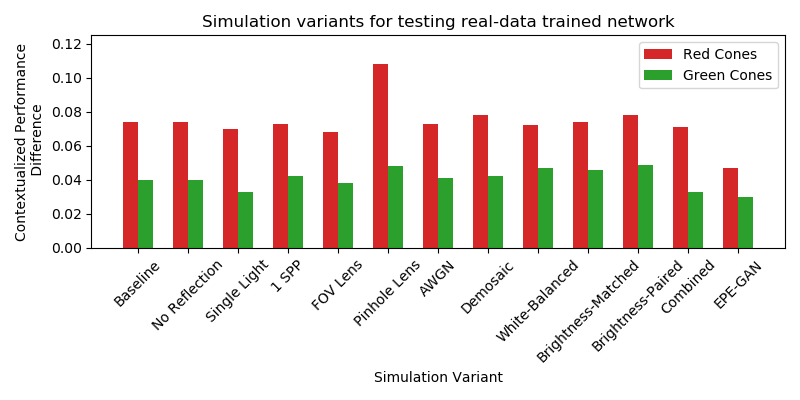}
	\caption{Predictive difference when evaluating a real-trained network in simulation. Lower is better.}
	\label{fig:cone_results:models_for_testing_real}
\end{figure}

First, we considered the indoor lab datasets to understand how {\SBELrealnet} performed on each simulation variant, and how similar that performance was to its performance on real data. To do this, we compared $real-sim_{variant}$ from the perspective of {\SBELrealnet} for each variant. The comparison was performed based on the contextualized performance methodology outlined in~\cite{elmquist2022performance}. Figure~\ref{fig:cone_results:models_for_testing_real} shows the sim-to-real difference for each of the simulation variants. The last variant (EPE-GAN) is a GAN-based augmentation described in~\cite{elmquist2022evaluating} which serves as a magnitude reference for a GAN approach to camera simulation enhancement. Additionally, we added a variant which combines multiple augmentations exhibiting a change from the baseline to understand how a combination of augmentations would perform. Specifically, the combined model included an FOV lens, single light, demosaicing, and brightness correction.

The results show that no model improved the simulated data as much as EPE-GAN. In fact, only a few of the augmentations changed the simulation significantly from this perspective. The largest change was when using a pinhole model, where the simulation became far less realistic, likely due to the significant difference in object shape. A few of the augmentations slightly improved the results, namely the FOV lens, using a single light, and the combined approach. However, these changes were minimal overall. The improvement for the FOV lens may be due to a narrowing of the cones, which could hide some minor inaccuracies in the 3D cone model.

\begin{figure}
	\centering
	\includegraphics[width=\linewidth]{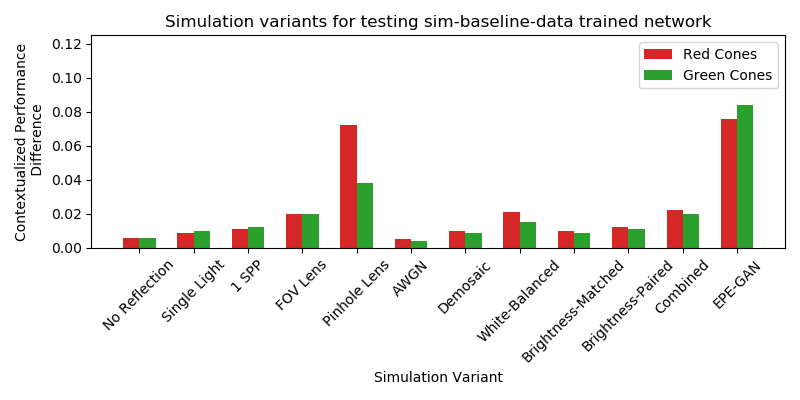}
	\caption{Predictive difference when evaluating a baseline sim-trained network in simulation. Shows the domain shift between the simulation variants. Lower performance difference means the simulations produce similar results.}
	\label{fig:cone_results:models_for_testing_sim}
\end{figure}

We conducted another study to understand the magnitude of the change from the baseline simulation by using each variant dataset to test the performance of a {\SBELsimnet}. These results are shown in Fig.~\ref{fig:cone_results:models_for_testing_sim}. Here, we see that most simulation variants induced a response in {\SBELsimnet} that was similar to the baseline dataset, except for two significant outliers. These outliers were EPE-GAN and pinhole. EPE-GAN results were as expected as this was the closest to reality, and introduced features and artifacts that may be difficult for a sim-trained network to detect; see~\cite{elmquist2022evaluating} for analysis and the related supplemental material~\cite{GANinSimSuplementalMaterial} for examples. The second outlier was the pinhole simulation, which suggests that shape was likely a factor in the learned policy in simulation. This was also expected due to the visually large differences in the pinhole example (Fig.~\ref{fig:cone:sim_pinhole_lens}).

While we begin to see which simulation variants are more realistic when testing, the greater opportunity for simulation is producing synthetic data for training new algorithms. To this end, we conducted a study of the cone variants where we trained variants of the object detector, one for each simulation variant (\SBELvarnet{variant}). With the trained model variant, we compared its performance on its own metric dataset with its performance on a reference metric dataset (real or baseline) to understand how well it predicted its own performance in those reference domains.

First, we looked at real data as the reference to understand how well the variant predicted its own performance in reality. For each variant, we measured the contextualized performance difference for each class. Figure~\ref{fig:cone_results:models_for_training_for_real} shows the magnitude of these differences. 
The results of EPE-GAN are the most predictive of the real performance, with no model variant having produced a network with such similar results. On the other hand, the white-balanced simulation results in poor predictive power, with the performance of {\SBELvarnet{white-balanced} on real being far different from its performance on itself and much worse than the baseline simulation. Each other variant was unremarkable relative to these two extremes. In absolute terms, none of these variants are satisfactorily close to reality; indeed, a difference between 0.1 and 0.3 (all other variants), when the full range of accuracy is defined 0-1, is insufficient to instill confidence in the sim-trained network. The last notable conclusion to draw from these results is that for each variant its network performed more similarly for green cones, except for the combined augmentations where the performance difference is almost identical. This observation motivates the need for future analysis.

\begin{figure}
	\centering
	\includegraphics[width=\linewidth]{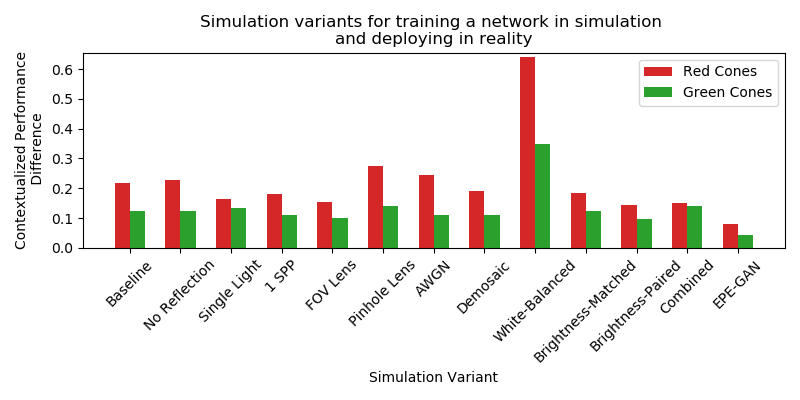}
	\caption{Predictive difference when training a model in sim intended for operation on real data. Lower difference is better. }
	\label{fig:cone_results:models_for_training_for_real}
\end{figure}

A similar study, where the baseline was used as the reference, was performed to understand how different these variants were from the baseline simulation. Results of this analysis are shown in Fig.~\ref{fig:cone_results:models_for_training_for_sim}. Many of the conclusions from the real-reference study are confirmed here, with white-balanced and EPE-GAN being among the largest differences. However, here we also see pinhole producing a network which does not work equivalently on the baseline. It should be noted that the magnitude for all sim variants vs. baseline are lower than that of the variant vs. real data, indicating that the effects of our augmentations on the whole are relatively minimal.

\begin{figure}
	\centering
	\includegraphics[width=\linewidth]{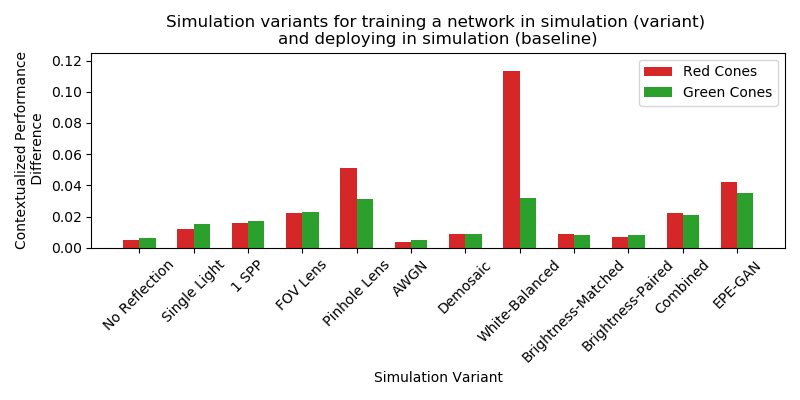}
	\caption{Predictive difference when training a model in sim intended for operation on baseline simulated data. Shows the domain shift between the simulation variants. Lower difference for similar models.}
	\label{fig:cone_results:models_for_training_for_sim}
\end{figure}

\subsection{City environment}

To understand the impact of the augmentations on GTAV data, we followed the same procedure as above by beginning with evaluating the performance of {\SBELcitynet} on each dataset. This study described how well each simulation variant predicted the performance of {\SBELcitynet} on real data. From previous work~\cite{elmquist2022evaluating}, we provide a reference of EPE-GAN to demonstrate the magnitude of a GAN-based approach to augmenting simulated data. The results from this study are provided in Fig.~\ref{fig:city_results:model_for_testing_real}. Evident in the results is that no augmentation performed as similar to reality as EPE-GAN. However, the white-balance match dataset is the closest; it is interesting to note however, that full brightness matching does not further improve the dataset. Both noise and distortion resulted in a dataset that was less predictive of reality. From a sensitivity perspective, distortion and color correction (apart from EPE-GAN) appeared to have the largest impact when evaluating {\SBELcitynet}.

\begin{figure}
	\centering
	\includegraphics[width=\linewidth]{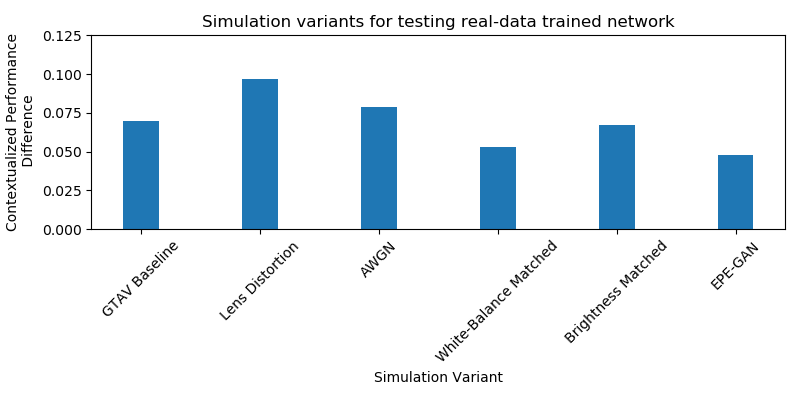}
	\caption{Predictive difference when evaluating a Cityscapes-trained network on GTAV variants. Lower is a variant producing results more similar to Cityscapes.}
	\label{fig:city_results:model_for_testing_real}
\end{figure}

We reran this comparison, looking at the performance of {\SBELgtanet} on each of the GTAV variants, to further understand how different the augmentations were from the baseline dataset. The results are shown in Fig.~\ref{fig:city_results:model_for_testing_sim}. Taking into account the scale of the plot, it is clear here that each augmentation (apart from GTAV-EPE) induced similar results to that in the baseline, with all differences below $0.02$. Only GTAV-EPE induced significantly different performance. This demonstrates the greater impact of scene changes such as the texture changes apparent in the GTAV-EPE images.

\begin{figure}
	\centering
	\includegraphics[width=\linewidth]{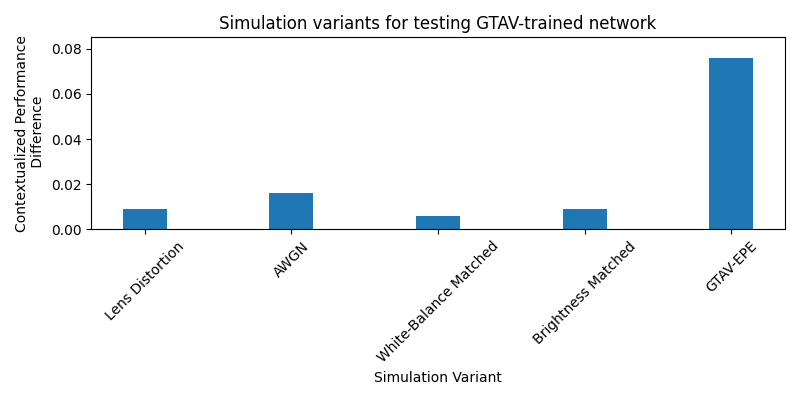}
	\caption{Predictive difference when evaluating a GTAV-trained network on GTAV variants. Lower is a variant more similar to baseline GTAV.}
	\label{fig:city_results:model_for_testing_sim}
\end{figure}

Finally, we looked at each GTAV variant for training a segmentation network. As training can expose additional biases in the data, this is important for understanding the effect of the augmentation. The results in Fig.~\ref{fig:city_results:model_for_training_for_real} include two reference marks: GTAV baseline and EPE-GAN, with EPE-GAN producing a network which performed most similar on its own and real data. As for the augmentations, we saw that lens distortion and noise had very low impact on the trained network. White-balance matching and brightness matching both significantly altered the trained network and in both cases the network's performance was less similar to reality. This is counter to the results for testing {\SBELcitynet} where these modifications slightly improved the GTAV data realism. This behavior, where augmenting a dataset can improve testing yet reduce the effectiveness of the dataset for training, is similar to previously reported results~\cite{elmquist2022evaluating} and is related to dataset variety and network robustness. 

\begin{figure}
	\centering
	\includegraphics[width=\linewidth]{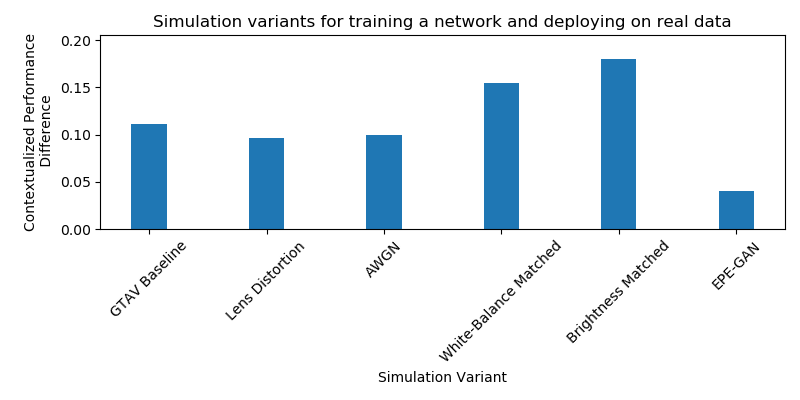}
	\caption{Predictive difference when evaluating a new network that was trained on the augmented data. Lower performance difference means the variant is more similar to baseline GTAV.}
	\label{fig:city_results:model_for_training_for_real}
\end{figure}

In all cases of applying AWGN with $\sigma$=0.01, the results were less significant than color or lens distortion augmentations. With a standard deviation of 0.01, we considered noise approximately 5 times that estimated in the real data. This indicates that, while it may have impact in specific scenarios or algorithms, noise could have been ignored for the two applications and algorithms included herein. Indeed, in these studies, it is unlikely realistic noise would alter the results with any significance.

\section{Conclusion and Future Work}
\label{sec:conclusion}

This contribution assesses the impact of including or excluding components of the camera model on the realism of a camera simulator, where realism is defined as the similarity of response manifested by a perception algorithm used by robots or autonomous vehicles. Specifically, we seek to understand where improvements to the usefulness of the images can be gained, and where additional modeling efforts are likely to produce little return on investment. 

The augmentations that produced significant change for our two perception algorithms in their respective environments were large-scale modifications to color and shape. We note that lens distortion can be significant as it induces structural changes which impact testing and training. Beyond detection, lens distortion could play a major role in any 3D projection of the image, where the location of pixels would determine 3D size or shape. Other large-scale changes such as color modification were significant factors in this study. Therefore, accurately modeling post-sensor image processing such as auto-contrast, white-balancing, and color correction would be important factors to include in closed-loop testing, particularly if few data augmentations were performed during training of the perception algorithm. 

However, none of the evaluated augmentations were as impactful as EPE-GAN, indicating that accurate modeling of the appearance of the virtual environment is critical. This is intuitive as white-balance and contrast are intertwined in the appearance of the virtual environment. As such, a hybrid approach, including precise modeling of distortion and post-sensor color adjustments, along with a GAN for scene-based enhancement could improve the quality of the camera simulators in robotics. These results quantitatively support other works where simulation of the environment played a critical role in the generalization of perception algorithms.

Small and fine-grain detail including noise, demosaicing, image quality, or slight changes to how exposure is modeled (not to be confused with \textit{if} exposure is modeled), resulted in very similar results to the baseline simulation. Important to this is the application and specific algorithms used, as different perception methods and tasks may be sensitive to other simulation phenomena. We argue that these should be well understood when leveraging simulation for training and testing of those algorithms and tasks.

These results indicate that focus and effort in simulation improvement should center on the environment and the large-scale phenomena in the camera model such as post-sensor color processing. Furthermore, modeling noise and demosaicing, or improving image quality may not provide significant value to our understanding of the object detection algorithm. This same procedure and methodology could be used to understand the impact of the camera component models for alternate perception tasks such as tracking or visual odometry, different perception algorithms/architectures, or other environments. When simulation is used for each of these tasks, the results may be different, with the model's importance determined in part by the task. In each case however, the approach will quantify the importance, giving insights into which aspects of simulation warrant improvement.

\section*{Acknowledgment}
This work was carried out in part with support from National Science Foundation project CPS1739869. Special thanks
to the ARC Lab at the University of Wisconsin-Madison for
their support through their motion capture facilities.

\bibliographystyle{IEEEtran}
\def\cprime{$'$}

\end{document}